\documentclass[journal]{IEEEtran}  % ,onecolumn
\usepackage{enumerate}
\usepackage{url}
\usepackage{flushend}
\usepackage{multirow}
\usepackage{colortbl}
\usepackage{pifont}
\usepackage{setspace}
\usepackage{cite}
\usepackage{algorithm}
\usepackage{algorithmic}
\usepackage[subfigure]{graphfig}
\usepackage{amsthm,amsmath,amssymb}
\usepackage{booktabs}
\usepackage{subfigure}
\usepackage{graphicx}
\usepackage{mathrsfs}
\usepackage{makecell}
\usepackage{cleveref}
\usepackage{footnote}
\usepackage{color}

\begin{document}
	%\let\cleardoublepage\clearpage
	%	\definecolor{ColorName}{rgb}{0,0.3,0.8} 
	\graphicspath{{simulationresults/}}

	\title{Continual learning-based probabilistic slow feature analysis for  monitoring multimode nonstationary processes}

		\author{Jingxin Zhang, Donghua Zhou, ~\IEEEmembership{Fellow,~IEEE},  Maoyin Chen, ~\IEEEmembership{Member,~IEEE}, and Xia Hong, ~\IEEEmembership{Senior Member,~IEEE}
			\thanks{This work was supported by National Natural Science Foundation of China [grant numbers 62033008, 61873143]. (Corresponding authors: Donghua Zhou; Maoyin Chen)}
			\thanks{Jingxin Zhang and Maoyin Chen are with the Department of Automation, Tsinghua University, Beijing 100084, China (e-mail: zjx18@mails.tsinghua.edu.cn; mychen@tsinghua.edu.cn). }
			\thanks{Donghua Zhou is with College of Electrical Engineering and Automation, Shandong University of Science and Technology, Qingdao 266000, China and also with the Department of Automation, Tsinghua University, Beijing 100084, China (e-mail: zdh@mail.tsinghua.edu.cn).}
			%	\thanks{Maoyin Chen is with the Department of Automation, Tsinghua University, Beijing 100001, China  (e-mail: mychen@tsinghua.edu.cn).}
			\thanks{Xia Hong is with Department of Computer Science, School of Mathematical, Physical and Computational Sciences, University of Reading, RG6 6AY, U.K.}
		}
	
	\maketitle
%	\IEEEpeerreviewmaketitle

	\begin{abstract}
	%	In this paper, a novel multimode nonstationary process monitoring approach is proposed by extending elastic weight consolidation (EWC) to  probabilistic slow feature analysis (PSFA) in order to extract multimode slow features for online monitoring. 
	This paper proposes a novel continual learning-based  probabilistic slow feature analysis (PSFA) algorithm  for monitoring multimode nonstationary processes, where multimode slow features are extracted and elastic weight consolidation (EWC) is adopted for sequential modes. 
	EWC  was originally introduced in the setting of machine learning of sequential multi-tasks with the aim of avoiding catastrophic forgetting issue, which equally poses as a major challenge in multimode nonstationary process monitoring.  When a new mode arrives,  a small set of data should be collected so that this mode can be identified by PSFA and limited prior knowledge. Then, a regularization term is introduced to  prevent new data from significantly interfering with the learned knowledge, where the parameter importance measures are estimated. The proposed method is denoted as PSFA--EWC, which  is updated continually and capable of achieving excellent performance. 
	%Different from traditional multimode  monitoring algorithms,
	 PSFA--EWC furnishes backward and forward transfer ability by a single model. The significant features of previous modes are retained while  consolidating new information, which  may contribute to learning new relevant modes. %Compared with several known methods, 
	The effectiveness  of the proposed method is demonstrated via a continuous stirred tank heater and a practical coal pulverizing system. 
	%	demonstrated
		%is employed to consolidating new information while retraining the learned features.
	\end{abstract}
	
	\def\abstractname{Note to Practitioners}
	\begin{abstract}
		Since industrial systems operate in varying modes and data are nonstationary within each mode, multimode nonstationary process monitoring is increasingly important. Traditional multimode monitoring methods generally need complete data from all possible modes and may need to be retrained from scratch when a new mode arrives, which require expensive computational resources and storage space. Besides, it is difficult to distinguish real faults from normal variations in multimode nonstationary processes. This paper proposes a novel continual learning-based probabilistic slow feature analysis, where  elastic weight consolidation  is employed to consolidate the previously learned knowledge while extracting multimode slow features. The monitoring model is updated sequentially and provides backward as well as forward transfer learning ability for successive modes. It is able to separate real faults from normal dynamics, which is beneficial to identifying a new mode for multimode nonstationary processes. In addition, the proposed approach delivers excellent model interpretability and  deals with missing data as well as uncertainty. In industrial applications, such as power plants and intelligent manufacturing processes, the proposed method can provide excellent monitoring performance.
	%	For practical industrial systems, such as large-scale power plants and intelligent manufacturing processes, the proposed method has outstanding ability to monitor various dynamic modes.
	\end{abstract}

	 
	% ,chen2022Data
	
	\section{Introduction}
	Data-driven process monitoring is vitally important for ensuring safety and reliability of modern industrial processes \cite{Rato2017translation,shi2020distributed,chen2022Data}.  Nonstationary process monitoring methods have been extensively studied \cite{qin2020bridging,jiang2018recursive,hu2020fault}. Slow feature analysis (SFA), which is effective in extracting invariant slow features from fast changing sensing data \cite{wiskott2002slow},  has been widely extended to process monitoring.  SFA could establish a comprehensive operating status, where the nominal operating deviations and real faults may be distinguished in the closed-loop systems \cite{shang2015concurrent,shang2018recursive,yu2019recursive,guo2016monitoring}.   
	%It has been mentioned that SFA-based methods could provide a comprehensive operating status, where the nominal operating deviations and the real faults may be distinguished in the closed-loop systems \cite{shang2015concurrent,shang2018recursive,yu2019recursive,shang2015probabilistic,guo2016monitoring}
	Recursive SFA (RSFA) \cite{shang2018recursive} and recursive exponential SFA \cite{yu2019recursive} were  developed and the associated parameters were updated for adaptive monitoring. Sufficient samples had been required to establish the initial model when a new mode was identified. Probabilistic SFA (PSFA) was proposed as a probabilistic framework
	% for nonstationary processes \cite{shang2015probabilistic,zafeiriou2016probabilistic} 
	with the advantage  of  effectively handling process noise  and uncertainties,  where measurement noise was modeled and missing data could  be settled conveniently \cite{guo2016monitoring}.

	Most industrial systems operate in multiple conditions due to equipment maintenance, market demands, changing of  raw materials, etc.  Multimode nonstationary process monitoring methods  have been investigated, which could be sorted into two categories \cite{quinones-grueiro2019data}, namely, single-model and multiple-model methods. Single-model methods transform the multimode data to  unimodal distribution \cite{ma2012a} or establish  adaptive models \cite{shang2018recursive,shang2018recursiveTCSA}. Local neighborhood standardization can normalize data into a single distribution and popular methods for one mode could be applied  \cite{ma2012a}.  However, the effectiveness may be influenced by the matching degree of  training and testing data.   
	%Recursive principal component analysis was adopted for adaptive process monitoring \cite{elshenawy2010efficient}.	
	Although  prior knowledge is not required,  these algorithms are effective for slow changing features and may fail to track the dramatic variations on the entire dataset \cite{shang2018recursive,ma2012a,shang2018recursiveTCSA}.
	
	The mainstream approaches of multimode monitoring are based on multiple-model schemes, where the modes are identified and  local models are built within each mode \cite{shao2020Bayesian,wen2015multimode,zhang2019an}. Mixture of canonical variate analysis (MCVA) was explored for multimode nonstationary processes \cite{wen2015multimode}.
	Improved mixture of probabilistic principal component analysis (IMPPCA) could be utilized for  multimode processes \cite{zhang2019an}, where the  model parameters  and the mode identification  were jointly optimized.  
	%$ K $-nearest neighbor is effective to classify the operating modes and then  local models were established \cite{ha2017multi}. 
   However, the number of modes is a priori and data from all possible modes are required before learning, which is infeasible and  time-consuming \cite{quinones-grueiro2019data}.  When  novel modes appear, sufficient data should be collected and new local models are relearned correspondingly. The model is only effective for the learned modes, but may be difficult to deliver excellent performance for similar modes \cite{zhang2019an,wen2015multimode}. Besides, multiple-model schemes may be redundant and difficult to identify modes accurately \cite{Huang2020structure}. The model's capacity and storage costs increase significantly with the emergence of modes.

%	Recently, the emergent research area of continual learning has received much attention in machine learning \cite{Kirkpatrick2017Overcoming,ven2020brain,Raia2020Embracing,MasseE10467Alleviating,rahaf2019continual}. It concerns the modeling of sequential tasks with the knowledge being acquired continually, fine-tuned and transferred throughout the entire learning process \cite{parisi2019continual}. Continual adaptation to the changing tasks is achieved by acquiring new information while preserving the learned knowledge. 	One long-standing challenge to be addressed is catastrophic forgetting issue, namely, learning a model with new information would influence the previously learned knowledge \cite{Kirkpatrick2017Overcoming}. Current continual learning-based algorithms focus on the image processing and generally require the class labels \cite{Kirkpatrick2017Overcoming,ven2020brain,Raia2020Embracing,rahaf2019continual,MasseE10467Alleviating,parisi2019continual}.   
	
	Recently, the emergent research area of continual learning 	
	has received much attention \cite{Kirkpatrick2017Overcoming,ven2020brain,Raia2020Embracing,MasseE10467Alleviating,parisi2019continual}.  One long-standing challenge to be addressed is catastrophic 	forgetting issue, namely, learning a model with new information 	would influence the previously learned knowledge \cite{Kirkpatrick2017Overcoming}. Continual learning is concerned with continual adaptation  of the model to the changing tasks   by acquiring new information while preserving the learned knowledge. While there are diverse techniques on continual learning ranging from  regularization \cite{Kirkpatrick2017Overcoming} to dynamic architectures \cite{MasseE10467Alleviating} to manipulating data memory replay \cite{ven2020brain},  the  majority benchmarking applications in the literatures appear  to   focus on the image processing and generally require the class labels \cite{Kirkpatrick2017Overcoming,ven2020brain,Raia2020Embracing,MasseE10467Alleviating,parisi2019continual,delange2021a}. Nevertheless, the concept of continual learning extends to lifelong machine learning \cite{chen2018Lifelong} as well as poses open problems to related areas of machine learning such as transfer learning \cite{Weiss2016A}, etc (The readers are referred to \cite{parisi2019continual} and references within). Of particular interest  here is its integration with domain-specific learning  such as autonomous agents \cite{parisi2019continual} and conditioning monitoring \cite{zhang2021multimode}.  
	One of the continual learning paradigms is called elastic weight consolidation (EWC) \cite{Kirkpatrick2017Overcoming},   in which it is analyzed that when a full data set of multiple tasks are decomposed based on  a sequence of incoming tasks, the  model parameters can be adjusted accordingly based on data from a new task, without sacrificing performance for any previously learned tasks. EWC was interpreted from Bayesian theory, thus providing excellent model interpretability \cite{Kirkpatrick2017Overcoming}.
	
	Similarly, in the context of multimode process monitoring, new modes would often appear continuously  and  different modes may share similar significant features \cite{Huang2020structure}.  In practical applications, it is often intractable to collect data from all modes. Zhang \textit{et al.} applied continual learning into multimode process monitoring \cite{zhang2021multimode}, where EWC was employed to settle the catastrophic forgetting of principal component analysis (PCA), referred to as PCA--EWC. However, data are assumed to be stationary in each mode and a mode is identified by statistical characteristics of data, which makes it ineffective for multimode nonstationary processes, as well as  difficult to distinguish the operating deviations and dynamic anomalies. Sometimes, it may be intractable to accurately identify the mode switching only by prior knowledge.

	Against this background, this paper considers a novel PSFA approach with continual learning ability, which is regarded as underlying multimode nonstationary processes for the observed sequential data.  
	%Data from each mode arrive sequentially and unknown modes are allowed. 
	%The important features of previous modes are retained in the model parameters when assimilating new information from novel modes, thus providing excellent performance for future relevant modes. 
	Moreover, the proposed algorithm would be best to distinguish  real faults and normal operating derivations. When a new mode is identified by PSFA and limited prior knowledge,  a small  set of data are collected before learning.   A quadratic penalty term is introduced to avoid the dramatic changes of mode-relevant parameters when a new mode is trained, where EWC is adopted to estimate the PSFA model parameter importance. PSFA assumes that the noise follows multivariate  Gaussian distribution in each mode, which makes it possible to estimate parameter importance by EWC. The proposed method is referred to as PSFA--EWC. 
	%The information from novel modes  is assimilated while consolidating the learned knowledge simultaneously, thus delivering continual learning ability for sequential modes.   

	The contributions  are summarized as follows:
	\begin{enumerate}[a)]
		\item PSFA with continual learning ability is firstly investigated for multimode nonstationary processes, where the mode  is identified by the  statistics and limited prior knowledge. PSFA--EWC  provides  excellent interpretability, and  deals with missing data, measurement noise  and uncertainty.
		\item The new information is extracted  while consolidating the previously learned  knowledge, which may aid the learning of future relevant modes and beneficial to monitoring the learned modes. Thus, PSFA--EWC furnishes the forward and backward transfer ability.
		
%		\item %Within the probabilistic framework, 
%		PSFA--EWC could provide  excellent interpretability, and  deal with missing data, measurement noise  and uncertainty.
	\end{enumerate}
	
	The rest of this paper is organized below.  Section \ref{sec2} reviews PSFA succinctly and outlines the  basic idea of our proposed approach. The technical core of PSFA--EWC  is detailed in Section \ref{sec3}. The monitoring procedure and  comparative experiments are designed in Section \ref{sec:monitoring}. The effectiveness of PSFA--EWC is illustrated by a continuous stirred  tank heater (CSTH) and a practical coal pulverizing system in Section \ref{sec4}. The conclusion is given in Section \ref{sec5}.

	\section{Preliminaries and problem statement}\label{sec2}
	For ease of exposition, we start with introducing the PSFA for a single mode, since it serves as basic ingredient of our proposed multimode PSFA. Then the basic idea of EWC as well as how to extend  EWC to multimode PSFA  is outlined. 
	
	\subsection{PSFA for a single mode}
	%In the probabilistic framework of  SFA, The objective is
	PSFA aims to identify the slowest varying latent features  from a sequence of time-varying observations ${\boldsymbol x_t}\in R^{m}$, $t= 1,2, \dots,T$,
	%Consider the observed data and the latent variables are Gaussian distributed, 
	which can be represented/generated via a state-space model with a first-order Markov chain architecture \cite{turner2007a}.   
	\begin{equation}\label{PSFA_model}
		\begin{aligned}
			{\boldsymbol x_t} = &\boldsymbol V{\boldsymbol y_t} + {\boldsymbol e_t},\quad {\boldsymbol e_t} \sim N\left( {\boldsymbol 0,\boldsymbol \Sigma_x } \right)\\
			{\boldsymbol y_t} =& \boldsymbol \Lambda {\boldsymbol y_{t - 1}} + {\boldsymbol w_t},\quad {\boldsymbol w_t} \sim N\left( {\boldsymbol 0,\boldsymbol \Sigma } \right)\\
			{\boldsymbol y_1} =& \boldsymbol u,\quad \boldsymbol u \sim N\left( {\boldsymbol 0,{\boldsymbol \Sigma _1}} \right)
		\end{aligned}
	\end{equation}
	where the low dimensional latent variable  ${\boldsymbol y_t} \in  R^{p}$, $p<m$.    $\boldsymbol \Lambda  = diag\left( {{\lambda _1}, \ldots ,{\lambda _p}} \right)$, with the constraint ${\boldsymbol \Lambda ^2} +\boldsymbol  \Sigma = \boldsymbol I$  to ensure the covariance matrix be the unit matrix 	$ \boldsymbol I $.
	%$ \boldsymbol I $ is an identity matrix with proper dimension. 
	The emission matrix is $ \boldsymbol V \in R^{m \times q} $ and measurement noise variance is $\boldsymbol \Sigma _x = diag\left( \sigma_1^2,\cdots,\sigma_m^2\right)$.
	
	For a single mode, the observed data and  latent slow features sequences are denoted as $ \boldsymbol X_s =\{{\boldsymbol x_t} \} \in R^{m \times T}$ and $ \boldsymbol Y_s =\{{\boldsymbol y_t}\}\in R^{p\times T}$, respectively.  $T$ is the number of samples and the estimation of $p $ has been discussed in \cite{guo2016monitoring}. 
	
	The joint distribution  is  given as \cite{zafeiriou2016probabilistic}
	\begin{equation}\label{PSFA-gen}
		%\begin{center}
		P\left( {\boldsymbol X_s|\boldsymbol Y_s} \right) = P\left( {{\boldsymbol y_1}} \right)\prod\limits_{t = 2}^T {P\left( {{\boldsymbol y_t}|{\boldsymbol y_{t - 1}}} \right)} \prod\limits_{t = 1}^T {P\left( {{\boldsymbol x_t}|{\boldsymbol y_t}} \right)} 
		%	\end{center}
	\end{equation}
	%	with the initial latent variables, the transition and emission distributions are
	%	\begin{equation}\label{y_initial}
	%		P\left( {{\boldsymbol y_1}} \right) = N\left( {{\boldsymbol y_1}|\boldsymbol 0,{\boldsymbol \Sigma _1}} \right)
	%	\end{equation}
	%	\begin{equation}\label{y_transition}
	%		P\left( {{\boldsymbol y_t}|{\boldsymbol y_{t - 1}}} \right) = N\left( {{\boldsymbol y_t}|\boldsymbol \Lambda {\boldsymbol y_{t - 1}},\boldsymbol \Sigma } \right)
	%	\end{equation}
	%	\begin{equation}\label{y_emission}
	%		P\left( {{\boldsymbol x_t}|{\boldsymbol y_t}} \right) = N\left( {{\boldsymbol x_t}|\boldsymbol V{\boldsymbol y_t},{\boldsymbol \Sigma _x}} \right)
	%	\end{equation}
	Let $ \theta_x = \left\{ \boldsymbol V,{\boldsymbol \Sigma _x}\right\}$, $ \theta_y = \left\{ {\boldsymbol \Sigma _1},\boldsymbol \Lambda \right\}$. The objective of PSFA  is to  estimate paramaters $\theta = \left\{\theta_x, \theta_y \right\} $ by maximizing the complete log likelihood function:
	%	$ p\left( {\boldsymbol X|\theta } \right) $  can be represented by
	%\begin{equation}\label{log-PSFA}
		\begin{align} \label{log-PSFA}
			\log \;P\left( {{\boldsymbol X_s},{\boldsymbol Y_s}|\theta } \right) 
			= &\sum\limits_{t = 1}^T {\log \;P\left( {{\boldsymbol x_t}|{\boldsymbol y_t},{\theta _x}} \right)}  + \log \;P\left( {{\boldsymbol y_1}|{\boldsymbol \Sigma _1}} \right) \nonumber \\
			&+ \sum\limits_{t = 2}^T {\log \;P\left( {{\boldsymbol y_t}|{\boldsymbol y_{t - 1}},\boldsymbol \Lambda } \right)} 
		\end{align}
%	\end{equation}
	The optimal parameter $\theta$ is optimized by maximizing \eqref{log-PSFA} using expectation maximization (EM) algorithm~ \cite{dempster1977maximum}.
%% (further details in Section \ref{sec3}). 	
	
%	According to (\ref{PSFA_model}), (\ref{log-PSFA}) is reformulated as
%	\begin{equation}\label{log_PSFA}
%		\begin{aligned} %\label{log_PSFA}
%			&\log \;P\left( {\boldsymbol X_s,{\boldsymbol Y_s}|\theta } \right) \\
%			%	= &  - \frac{{m + p}}{2}T\log \;2\pi   - \frac{{T - 1}}{2}\log \left| \boldsymbol \Sigma  \right| - \frac{1}{2} \boldsymbol y_1^T \boldsymbol \Sigma _1^{ - 1}{\boldsymbol y_1}\\
%			= & - \frac{1}{2} \left\{ ({m + p})T\log \;2\pi  +(T-1)\log \left| \boldsymbol \Sigma  \right| +\boldsymbol y_1^T \boldsymbol \Sigma _1^{ - 1}{\boldsymbol y_1} \right. \\
%			&\left. +{T\log \left| {{\boldsymbol \Sigma _x}} \right| + \sum\limits_{t = 1}^T {{{\left( {{\boldsymbol x_t} -\boldsymbol V{\boldsymbol y_t}} \right)}^T} \boldsymbol \Sigma _x^{ - 1}\left( {{\boldsymbol x_t} -\boldsymbol V{\boldsymbol y_t}} \right)} } \right.\\
%			& \left. +\log \left| {{\boldsymbol \Sigma _1}} \right| + { { \sum\limits_{i = 2}^T {{{\left( {{\boldsymbol y_i} - \boldsymbol \Lambda {\boldsymbol y_{i - 1}}} \right)}^T}{\boldsymbol \Sigma ^{ - 1}}\left( {{\boldsymbol y_i} -\boldsymbol \Lambda {\boldsymbol y_{i - 1}}} \right)} } } \right\} 
%		\end{aligned}
%	\end{equation}

%	
	%\subsection{Elastic weight consolidation for multimode PSFA}
	\subsection{Problem statement}
%	We explain the basic idea of extending	EWC for multimode PSFA processes and then summarize the key objectives of proposed PSFA--EWC. 
	The problem is interpreted  for multimode nonstationary processes and then the  key objective is summarized.
	Consider also based on PSFA model \eqref{PSFA_model}, in the multimode scenario where data stream are generated as incoming new modes  $\mathcal{M}_K$, $K=1,2,\dots$ one at time. For each mode $\mathcal{M}_K$, normal data $\boldsymbol X_K \in R^{m \times T_K}$ are collected, where $ T_K $  is the number of samples. Correspondingly it is assumed that $\boldsymbol Y_K \in R^{p\times T_K}$ need to be extracted from the $K${th} mode. Denote the total observed data and its latent slow features as $\boldsymbol X =\{\boldsymbol X_1, \boldsymbol X_2, \dots \}$, $\boldsymbol Y =\{\boldsymbol Y_1, \boldsymbol Y_2, \dots \}$. 
	
	%the use of
	EWC initially considers Bayesian rule for the sequential learning process in which the most probable parameters should be found by maximizing the  conditional probability  \cite{Kirkpatrick2017Overcoming}
	%. Based on  Bayesian rule, the conditional probability is calculated by:
	\begin{equation}\label{bayesian1}
		\log \,P\left( {\theta |\boldsymbol X ,\boldsymbol Y} \right) = \log \,P\left( {\boldsymbol X, \boldsymbol Y |\theta } \right) + \log \,P\left( \theta  \right) - \log \,P\left( \boldsymbol X, \boldsymbol Y \right)
	\end{equation}
	where  $P(\theta)$ is prior probability and  $P(\boldsymbol X, \boldsymbol Y|\theta)$  is the data probability. For illustration only the first two successive independent modes ${\mathcal{M}_1}$ and ${\mathcal{M}_2}$ are initially considered.
	% mode ${\mathcal{M}_1}$ ($\boldsymbol X_1$) and  mode ${\mathcal{M}_2}$ ($\boldsymbol X_2$). 
	Then, \eqref{bayesian1} can be reformulated as \cite{Kirkpatrick2017Overcoming}:
	\begin{align}\label{poster2}
		\log \,P\left( {\theta |\boldsymbol X, \boldsymbol Y } \right) = &\log \,P\left( {\boldsymbol X_2,\boldsymbol Y_2 |\theta } \right) + \log \,P\left( \theta |\boldsymbol X_1, \boldsymbol Y_1 \right)\nonumber\\
		&  + don't \ care \ terms
		%- \log \,P\left( \boldsymbol X_2 \boldsymbol Y_2\right)
	\end{align}
	where $P\left( {\theta |\boldsymbol X, \boldsymbol Y} \right)  $ is the posterior probability of the parameter given the entire dataset. $P\left( {\boldsymbol X_2,\boldsymbol Y_2 |\theta } \right)$ represents the loss function for mode $ {\mathcal{M}_2} $. Posterior distribution $P\left( \theta |\boldsymbol X_1,\boldsymbol Y_1  \right) $  can reflect all information of mode $ {\mathcal{M}_1} $ \cite{Kirkpatrick2017Overcoming}. This equation reflects the key idea of EWC in continual learning framework of updating system parameters based on a composite cost function that is dependent on current parameters learned from previous data and new incoming data by using posterior distribution $P\left( \theta |\boldsymbol X_1,\boldsymbol Y_1  \right) $  which acts as a constraint in future objective, so that the learned knowledge will not be forgotten. 
	
	\begin{figure}[!tbp]
		\centering
		\includegraphics[width=0.40\textwidth]{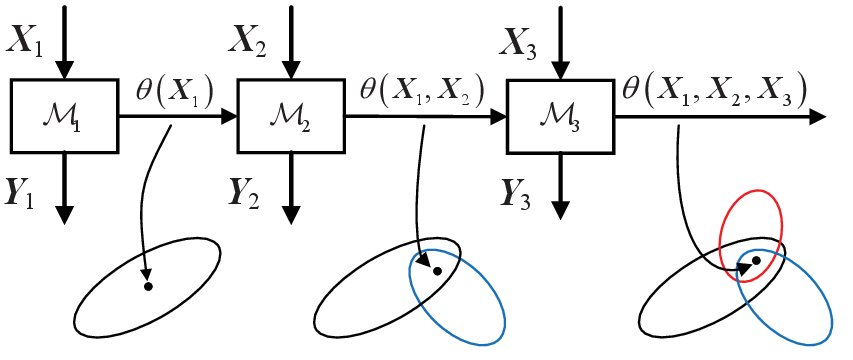}
		\caption{An illustration of the proposed PSFA--EWC with continual learning ability for three consecutive modes $\mathcal{M}_1$, $\mathcal{M}_2$, $\mathcal{M}_3$. }
		%The multimode slow features for each mode are extracted, while the model parameter $\theta$ is continually updated using only data of a new mode, while maintaining performance of all old modes.  Black circle: optimal parameter region that log likelihood for mode $\mathcal{M}_1$ is maximized; Blue circle: optimal parameter region that log likelihood for mode $\mathcal{M}_2$ is maximized; Red circle: optimal parameter region that log likelihood for mode $\mathcal{M}_3$ is maximized.}
		\label{Illustra-PSFA-EWC}
	\end{figure}

	This is the first time that continual learning-based PSFA is proposed for monitoring where new optimization procedures of  PSFA--EWC   will be introduced in Section \ref{sec3}, as depicted in Fig. \ref{Illustra-PSFA-EWC} for three modes. The multimode slow features for each mode are extracted, while the  parameter $\theta$ is continually updated using only data of a new mode, while maintaining performance of all old modes.  
	Black, blue and red circles represent the optimal parameter regions that the log likelihoods for modes  $\mathcal{M}_1$, $\mathcal{M}_2$  and $\mathcal{M}_3$ are maximized, respectively.
	%Black circle represents the optimal parameter region that log likelihood for mode $\mathcal{M}_1$ is maximized. Blue circle is the optimal parameter region that log likelihood for mode $\mathcal{M}_2$ is maximized. Red circle denotes the optimal parameter region that log likelihood for mode $\mathcal{M}_3$ is maximized.	
	This process can be generalized to $K >3$ modes, with
%	\begin{align}\label{poster2}
%		& \log \,P\left( {\theta |\boldsymbol X, \boldsymbol Y } \right) = \log \,P\left( {\boldsymbol X_K,\boldsymbol Y_K |\theta } \right)   \nonumber\\
%		& \ \ \ \ \ + \log \, P\left( \theta |{\cal M} _{i=1}^{K-1}  \right)+ don't \ care \ terms
%		%- \log \,P\left( \boldsymbol X_2 \boldsymbol Y_2\right)
%	\end{align}
%\begin{equation}
	\begin{align}\label{poster2}
	 \log \,P\left( {\theta |\boldsymbol X, \boldsymbol Y } \right) 
	= &\log \,P\left( {\boldsymbol X_K,\boldsymbol Y_K |\theta } \right) + \log \,   P\left( \theta |{\cal M} _{i=1}^{K-1}  \right) \nonumber \\ 
	 & + don't \ care \ terms
	%- \log \,P\left( \boldsymbol X_2 \boldsymbol Y_2\right)
\end{align}
%\end{equation}
	where 
	$P\left( \theta |{\cal M} _{i=1}^{K-1}  \right) \triangleq P\left( \theta |\boldsymbol X_1, ...,\boldsymbol X_{K-1},\boldsymbol Y_1,   ..., \boldsymbol Y_{K-1 } \right)$.
	
	The first term in \eqref{poster2} is complete likelihood for $K$th mode. The  second term in \eqref{poster2} is parameter estimate that reflects information from all previous modes, thus can be interpreted as log prior probability of parameter for $K$th mode. Since it is assumed that data from all previous modes will not be accessible
	to obtain $P\left( \theta |{\cal M}_{i=1}^{K-1}  \right)$ exactly, it is found by recursive  approximation as detailed in Section \ref{sec3:1}.
	% and significant information of mode $ {\mathcal{M}_1} $ is contained in the posterior probability.
	% (see (\ref{log-PSFA}))

	\section{The proposed PSFA--EWC algorithm}\label{sec3}
	
	\subsection{Recursive Laplacian approximation of $P\left( \theta |{\cal M} _{i=1}^{K-1}  \right)$}\label{sec3:1}
	Consider the multimode PSFA process  where data are collected sequentially with mode index $K=1,2,3, ...$. For the sake of notational simplicity,   it is assumed in the sequel that the data  ${\boldsymbol x_t}$ and corresponding slow features ${\boldsymbol y_t}$ start from $t=1$ at the beginning and end at $t=T_K$ of the $K$th mode.  The proposed PSFA--EWC algorithm starts with solving an initial single mode model as $K=1$. 
	%It is initially assumed that 
	An optimal parameter, denoted as $ \theta _{{\mathcal{M}_1}}^* $, has been obtained from the first mode  $\mathcal{M}_1$ based on solving \eqref{log-PSFA}.  For later modes ($K  \ge  2$), the monitoring model is updated recursively based on the data from $K$th mode  and the current monitoring model before $K$, where EM is employed \cite{dempster1977maximum} to solve the optimization problem of maximizing $J(\theta)$ in Section \ref{sec3:2}.  
	%	acquired by 
	%	\begin{equation}
	%		\theta _{{\mathcal{M}_1}}^* = \arg \max_{\theta} \; \log P(\theta|\boldsymbol X_1)
	%	\end{equation}

%	Specifically consider initially the case of two modes $K=2$,
	Consider initially two modes $K=2$,
	 %then the term 
	 $\log P(\theta|{\cal M}_1)  $ in \eqref{poster2} is approximated by the Laplace approximation \cite{Kirkpatrick2017Overcoming}  as
	\begin{equation}\label{secondtaylor}
		\begin{aligned}
			\log P(\theta|{\cal M}_1) 
			\approx &- \frac{1}{2}(\theta-{\theta _{{\mathcal{M}_1}}^*})^T \left(T_1 \boldsymbol F ({\theta _{{\mathcal{M}_1}}^*})+ \lambda_{prior}\boldsymbol I \right)\\
			& \cdot(\theta-{\theta _{{\mathcal{M}_1}}^*}) + constant\nonumber\\
		%	=& - \frac{1}{2}(\theta-{\theta _{{\mathcal{M}_{1}}}^*})^T 			{\boldsymbol \Omega}_{\mathcal{M}_{1}} (\theta-{\theta _{{\mathcal{M}_{1}}}^*}) + constant
		\end{aligned}
	\end{equation}
 where  $\boldsymbol F({\theta _{{\mathcal{M}_1}}^*}) $ is Fisher information matrix (FIM) and computed by \eqref{estimation_FIM} in Appendix \ref{appendix:FIM}.  $\lambda_{prior} \boldsymbol I $ is the Guassian prior precision matrix for mode $ {\mathcal{M}_1} $. %However, the sample size would have non-negligible influence on the approximation. To ensure the approximation quality, a mode-specific hyperparameter $\eta_{1} >0$ is introduced to replace  $T_1$ \cite{huszar2017on}, namely,
 The sample size $T_1$ may have non-negligible influence on the approximation, which would be replaced by a mode-specific hyperparameter $\eta_{1} >0$   to enhance the approximation quality \cite{huszar2017on}, namely,
\begin{equation}
	\log P(\theta|{\cal M}_1) = - \frac{1}{2}(\theta-{\theta _{{\mathcal{M}_{1}}}^*})^T
	{\boldsymbol \Omega}_{\mathcal{M}_{1}} (\theta-{\theta _{{\mathcal{M}_{1}}}^*}) + constant \nonumber
\end{equation}
	where   ${\boldsymbol \Omega}_{\mathcal{M}_1}  = \eta_{1}  \boldsymbol F({\theta _{{\mathcal{M}_1}}^*})+ \lambda_{prior}\boldsymbol I$.
	% $\boldsymbol F({\theta _{{\mathcal{M}_1}}^*}) $ is the Fisher information matrix (FIM) and computed by (\ref{estimation_FIM}). %  in Appendix \ref{appendix:FIM}.  
	%To avoid the influence of sample size and  better control over the approximation, a mode-specific hyperparameter $\eta_{1} >0$ is introduced to replace  $T_1$ \cite{huszar2017on}.  $\lambda_{prior} \boldsymbol I $ is the Guassian prior precision matrix for mode $ {\mathcal{M}_1} $.
	%$\eta_1>0$ are $\lambda_{prior}>0$ are regularization parameters.
	%we consider the successive independent modes and the data are collected sequentially. 
	
	%in $J(\theta)$
	When the $K$th mode $\mathcal{M}_K$ arrives ($K \ge 3$), we approximate $\log P\left( \theta |{\cal M} _{i=1}^{K-1}  \right)$    by recursive Laplace approximation as
	\begin{equation}\label{moremode}
		\begin{aligned}
			\log P(\theta|{\cal M} _{i=1}^{K-1} ) 
			\approx &- \frac{1}{2}(\theta-{\theta _{{\mathcal{M}_{K-1}}}^*})^T
			{\boldsymbol \Omega}_{\mathcal{M}_{K-1}}(\theta-{\theta _{{\mathcal{M}_{K-1}}}^*})  \nonumber\\
			&+ constant
		\end{aligned}
	\end{equation}
	where
	\begin{equation}\label{importance_update}
		{\boldsymbol \Omega}_{\mathcal{M}_{K-1}} = {\boldsymbol \Omega}_{\mathcal{M}_{K-2}}+ \eta_{{K-1}} \boldsymbol F_{\mathcal{M}_{K-1}},  \ \ \  K \geq 3
	\end{equation}
 $ \boldsymbol F_{\mathcal{M}_{K-1}} $ is FIM of mode $\mathcal{M}_{K-1}$ and $ \eta_{K-1} $ is a hyperparameter.   $\log \; P\left( \theta |{\cal M} _{i=1}^{K-1}  \right)$ is approximated by a quadratic term centered at current optimum, with %weighting 
 ${\boldsymbol \Omega}_{\mathcal{M}_{K-1}}$ acting as an importance measure regulating data from $K-1$ modes.

	\subsection{PSFA--EWC algorithm}\label{sec3:2}
	Consider the objective of PSFA--EWC of  maximizing
	\begin{equation}\label{psfaewcaim}
		J\left(\theta \right) 
		= \log \,P\left( {\boldsymbol X_K,\boldsymbol Y_K|\theta } \right) + \log P\left( \theta |{\cal M} _{i=1}^{K-1}  \right)
	\end{equation}
	subject to PFSA model \eqref{PSFA_model}.
	%Within the PSFA framework, the latent slow features are denoted as $ \boldsymbol Y_K $. 
	Recall \eqref{log-PSFA}, the log-likelihood function for the current mode $\mathcal{M}_K$ is represented by
	%	$\log \; p\left( {\boldsymbol X_K|\theta } \right) $  can be represented by
	%\begin{equation}\label{PSFA_single}
		\begin{align} \label{PSFA_single}
			\log \;P\left( {{\boldsymbol X_K},{\boldsymbol Y_K}|\theta } \right) %\\
			= &\sum\limits_{t = 1}^{T_K} {\log \;P\left( {{\boldsymbol x_t}|{\boldsymbol y_t},{\theta _x}} \right)}  + \log \;P\left( {{\boldsymbol y_1}|{\boldsymbol \Sigma _1}} \right) \nonumber\\
			&+ \sum\limits_{t = 2}^{T_K} {\log \;P\left( {{\boldsymbol y_t}|{\boldsymbol y_{t - 1}},\boldsymbol \Lambda } \right)}
		\end{align} 
%	\end{equation}
	
	The regularization term is designed as
	\begin{equation}\label{regu_term}
		\begin{aligned}
			\log P\left( \theta |{\cal M} _{i=1}^{K-1}  \right) 
			\approx &  -\gamma_{1,K} \left\|\boldsymbol V - \boldsymbol V_{\mathcal{M}_{K-1}} \right\|_{{\boldsymbol \Omega}_{\mathcal{M}_{K-1}}^V}^2 \\
			&- \gamma_{2,K}  \sum\limits_{i = 1}^p {{\Omega}_{\mathcal{M}_{K-1},i}^{\lambda} {{\left( {{\lambda _i} - {\lambda _{\mathcal{M}_{K-1},i}}} \right)}^2}}
		\end{aligned}   
	\end{equation}
	where ${{\boldsymbol \Omega}_{\mathcal{M}_{K-1}}^V}$ and ${\Omega}_{\mathcal{M}_{K-1},i}^{\lambda}$  measure the importance of $ \boldsymbol V_{\mathcal{M}_{K-1}} $ and $\lambda _{\mathcal{M}_{K-1},i}$,  $i=1, \cdots, p$. ${ \lambda}_{\mathcal{M}_{K-1},i}$ and ${\Omega}_{\mathcal{M}_{K-1},i}^{\lambda}$ are the $i$th elements of diagonal matrices $\boldsymbol \Lambda_{\mathcal{M}_{K}}$ and ${\boldsymbol \Omega}_{\mathcal{M}_{K-1}}^{\Lambda}$, which are the optimal parameters of mode $ \mathcal{M}_{K-1} $.  $ \gamma_{1,K} $ and $ \gamma_{2,K} $ are user-defined hyperparameters.  The setting $ \gamma_{1,K} $ and $ \gamma_{2,K} $ makes it flexible to adjust the weights of previous modes. 

%
%%	
	
%	\textcolor{red}{The setting $ \gamma_{1,K} $ and $ \gamma_{2,K} $ allows users to obtain models with more focus on particular mode, less on others .. }
%	\textcolor{red}{The difference $ \gamma_{1,K} $ and $ \gamma_{2,K} $ to $\eta$ are ... }	
	
	%They are positive definite symmetric and updated by (\ref{importance_update}). 
	% that measures the importance of previous $ K-1 $  modes.	
	For  the proposed PSFA--EWC, the total objective function of $K$ modes  can be formally described by
%	\begin{equation}\label{PSFAWC_aim}
		\begin{align} \label{PSFAWC_aim} 
			J\left(\theta \right) 
			%= &\log \,P\left( {\boldsymbol X_K|\theta } \right) + \log \,P\left( \theta |\boldsymbol X_1,\cdots,\boldsymbol X_{K-1} \right)\\
			=  &\sum\limits_{t = 1}^{T_K} {\log \;P\left( {{\boldsymbol x_t}|{\boldsymbol y_t},{\theta _x}} \right)}  	+ \sum\limits_{t = 2}^{T_K} {\log \;P\left( {{\boldsymbol y_t}|{\boldsymbol y_{t - 1}},\boldsymbol \Lambda } \right)} \nonumber \\ 
			&+ \log \;P\left( {{\boldsymbol y_1}|{\boldsymbol \Sigma _1}} \right) 
			%	&	+ \sum\limits_{t = 2}^T {\log \;P\left( {{\boldsymbol y_t}|{\boldsymbol y_{t - 1}},\boldsymbol \Lambda } \right)}\\ 
			-\gamma_{1,K} \left\|\boldsymbol V - \boldsymbol V_{\mathcal{M}_{K-1}} \right\|_{{\boldsymbol \Omega}_{\mathcal{M}_{K-1}}^V}^2 \nonumber \\
			&  - \gamma_{2,K}  \sum\limits_{i = 1}^p {{\Omega}_{\mathcal{M}_{K-1},i}^{\lambda} {{\left( {{\lambda _i} - {\lambda _{\mathcal{M}_{K-1},i}}} \right)}^2}} 
		\end{align}
%	\end{equation}
	subject to 	the PSFA model \eqref{PSFA_model}.
	Note that for $K>2$, since the quadratic penalty is added, it slows down the changes to parameters with respect to the previously optimum values that are obtained in learned modes \cite{parisi2019continual,ven2020brain}. In other words, the parameters that result in significant deterioration in performance of previous modes will be penalized, avoiding catastrophic forgetting problem.
	%	\subsubsection{EM for solution}
	
	When $K=1$, ${{\boldsymbol \Omega}_{\mathcal{M}_{K-1}}^V} = \boldsymbol 0$, ${\boldsymbol \Omega}_{\mathcal{M}_{K-1}}^{\Lambda} = \boldsymbol 0$. There is no need to provide $ \boldsymbol V_{\mathcal{M}_{K-1}} $ and $ {\boldsymbol \Lambda _{\mathcal{M}_{K-1}}} $, this means that the proposed PSFA--EWC algorithm has a unified formulation as a sequential single mode based on $K$th mode data only, with current parameters used as quadratic penalty, which are updated via recursive Laplacian approximation between each mode in Section  \ref{sec3:1}. The EM  \cite{dempster1977maximum}  is employed to optimize the parameter $\theta  = \left\{ {\boldsymbol V, \boldsymbol \Lambda,  {\boldsymbol \Sigma _x}, {\boldsymbol \Sigma _1}} \right\}$ by solving \eqref{PSFAWC_aim}. 
	
%	\subsubsection*{\textbf{EM algorithm for solving (\ref{PSFAWC_aim})}}	
	
	\subsubsection{\textbf{E-step}}	
	%The traditional calculation of sufficient statstics are 
	%Three sufficient statistics are required to be available for M-step, namely, 
	Assume that $\theta$ is available, the	E-step  estimates three sufficient statistics, namely,
	% $\mathbb E \left[\boldsymbol y_t|\boldsymbol X_K\right]$, $\mathbb E \left[\boldsymbol y_t \boldsymbol y_{t-1}|\boldsymbol X_K\right]$ and $\mathbb E \left[\boldsymbol y_t \boldsymbol y_t^T|\boldsymbol X_K\right]$. 	
\begin{equation}\label{mean}
		 \mathbb E \left[\boldsymbol y_t|\boldsymbol X_K\right] =  \hat{\boldsymbol \mu}_t 
\end{equation}			
\begin{equation}\label{covariance}
	 \mathbb E \left[\boldsymbol y_t \boldsymbol y_{t-1}^T|\boldsymbol X_K\right] =  \boldsymbol J_{t-1} \hat{\boldsymbol U}_t +\hat{\boldsymbol \mu}_t \hat{\boldsymbol \mu}_{t-1}^T 	
\end{equation}		
\begin{equation}\label{variance}
	  \mathbb E \left[\boldsymbol y_t \boldsymbol y_t^T|\boldsymbol X_K\right] = \hat{\boldsymbol U}_t+\hat{\boldsymbol \mu}_t \hat{\boldsymbol \mu}_{t}^T
\end{equation}
Detailed information has been summarized in Appendix \ref{sufficient}.	
	 %to calculate statistics, 
%		Similar to \cite{zafeiriou2016probabilistic}, Kalman filter  and Tanch-Tung-Striebel (RTS) smoother \cite{sarkka2008unscented} are adopted,  which contains the forward and backward recursion steps. 
%	
%	First,  the forward recursions are adopted to  estimate the posterior distribution
%	$P\left( {{\boldsymbol  y_t}|{\boldsymbol  x_1},{\boldsymbol  x_2}, \cdots ,{\boldsymbol  x_t},{\theta ^{old}}} \right) \sim N\left( {{\boldsymbol  \mu _t},{\boldsymbol  U_t}} \right)$ sequentially. The posterior marginal is calculated by
%	%\begin{equation}
%	\begin{center}
%		$		\begin{aligned}
%			&\int {N\left( {{\boldsymbol y_{t - 1}}|{\boldsymbol \mu _{t - 1}},{\boldsymbol U_{t - 1}}} \right)} N\left( {{\boldsymbol y_t}|\boldsymbol \Lambda {\boldsymbol y_{t - 1}},\boldsymbol \Sigma } \right)d{\boldsymbol y_{t - 1}} \\
%			&= N\left( {{\boldsymbol y_t}|\boldsymbol \Lambda {\boldsymbol y_{t - 1}},{\boldsymbol P_{t - 1}}} \right)
%		\end{aligned}$
%	\end{center}
%	%	\end{equation}
%	where $\boldsymbol P_{t-1}$ is the variance.
%	%	which needs to be evaluated in order to obtain the posterior marginal. 
%	
%	Then, parameters of the posterior distribution $P\left( \boldsymbol Y_K| \boldsymbol X_K, \theta^{old}\right)$ are acquired by backward recursion steps. The procedure is summarized in Algorithm \ref{alg:estep} Appendix.
%	%\ref{alg:estep}.
%	

	\subsubsection{\textbf{M-step}}	
	%	The model parameters $\theta  = \left\{ {\boldsymbol V,{\boldsymbol \Sigma _x},{\boldsymbol \Sigma _1},\boldsymbol \Lambda } \right\}$ are available when calculating the sufficient statistics. Then,
	
	Assume that three sufficient statistics are fixed, the parameters are updated alternately. % to optimize  (\ref{PSFAWC_aim}). 
	
	Since $ \boldsymbol V $ and $ \boldsymbol \Sigma _x $ are contained in $ P\left( {{\boldsymbol x_t},{\boldsymbol y_t}|{\theta _x}} \right) $ and the regularization term $ {\gamma _{1,K}}\left\| {\boldsymbol V - {\boldsymbol V_{\mathcal{M}_{K-1}}}} \right\|_{{\boldsymbol \Omega}_{\mathcal{M}_{K-1}}^V} ^2 $, then 
	%	According (\ref{Qfunction}) and (\ref{PSFAWC_aim}), 
	\begin{equation}
		\left\{ {{\boldsymbol V^{new}},\boldsymbol \Sigma _x^{new}} \right\}
		= \arg \mathop {\max }\limits_{\boldsymbol V,{\boldsymbol \Sigma _x}} \; \mathop J\left( {\boldsymbol V,{\boldsymbol \Sigma _x}} \right)
	\end{equation}
	where 
	%\begin{equation}
	\begin{center}
		$	\begin{aligned}
			&J\left( {\boldsymbol V,{\boldsymbol \Sigma _x}} \right)\\
			= & \sum\limits_{t = 1}^{T_K} {\log \;P\left( {{\boldsymbol x_t},{\boldsymbol y_t}|{\theta _x}} \right)} -{\gamma _{1,K}}\left\| {\boldsymbol V - {\boldsymbol V_{\mathcal{M}_{K-1}}}} \right\|_{{\boldsymbol \Omega}_{\mathcal{M}_{K-1}}^V} ^2 \\
			=& - \frac{T_K}{2}\log \left| {{\boldsymbol \Sigma _x}} \right|	- \frac{1}{2}\sum\limits_{t = 1}^{T_K} \left(  
			tr\left( {\mathbb{E}\left[ {{\boldsymbol y_t}\boldsymbol y_t^T| \boldsymbol X_K} \right]} \right){\boldsymbol V^T} \boldsymbol \Sigma _x^{ - 1}\boldsymbol V \right.\\
			& \left. +tr\left( {{\boldsymbol x_t} \boldsymbol x_t^T \boldsymbol \Sigma _x^{ - 1}} \right)   -2tr\left( \boldsymbol x_t^T\boldsymbol \Sigma _x^{ - 1} \boldsymbol V \mathbb{E}\left[ {{\boldsymbol y_t}|\boldsymbol X_K} \right] \right) \right) \\
			& -	{\gamma _{1,K}}tr\left( {{(\boldsymbol V - {\boldsymbol V_{\mathcal{M}_{K-1}}})^T}{{\boldsymbol \Omega}_{\mathcal{M}_{K-1}}^V} (\boldsymbol V - {\boldsymbol V_{\mathcal{M}_{K-1}}})} \right) 
		\end{aligned}$
	\end{center}
	Let the derivative with respect to  $\boldsymbol V $  be zero, then
	\begin{equation}\label{first_dev_v}
		\begin{aligned}
			%			&\sum\limits_{t = 1}^T {\boldsymbol \Sigma _x^{ - 1}\boldsymbol x_t \mathbb E\left[ {\boldsymbol y_t^T|\boldsymbol X} \right]}  - \sum\limits_{t = 1}^T {\boldsymbol \Sigma _x^{ - 1}\boldsymbol V \mathbb E\left[ {{\boldsymbol y_t}\boldsymbol y_t^T|\boldsymbol X} \right]}  - {\gamma _1}{{\boldsymbol \Omega}_{\mathcal{M}_{K-1}}^V}\left( {\boldsymbol V - {\boldsymbol V_{\mathcal{M}_{K-1}}}} \right) = 0\\
			%			\Rightarrow &
			&\sum\limits_{t = 1}^{T_K} {\boldsymbol x_t \mathbb E\left[ {\boldsymbol y_t^T|\boldsymbol X_K} \right]}  + {\gamma _{1,K}}{\boldsymbol \Sigma _x}{{\boldsymbol \Omega}_{\mathcal{M}_{K-1}}^V} {{\boldsymbol V_{\mathcal{M}_{K-1}}}} \\
			&= \boldsymbol V\sum\limits_{t = 1}^{T_K} {\mathbb E\left[ {{\boldsymbol y_t}\boldsymbol y_t^T|\boldsymbol X_K} \right]}  + {\gamma _{1,K}}{\boldsymbol \Sigma _x} {{\boldsymbol \Omega}_{\mathcal{M}_{K-1}}^V} \boldsymbol  V
		\end{aligned} 
	\end{equation}
	This problem is actually the Sylvester equation and the solution is denoted as $ \boldsymbol V^{new} $.

	Taking the derivative about $ {\sigma _i^2} $ and let it be zero, then 
%	\begin{equation}\label{sigma_x}
		\begin{align} \label{sigma_x}
			{\left( {\sigma _i^2} \right)^{new}} =& \frac{1}{T_K}\sum\limits_{t = 1}^{T_K} \left\{ \mathbb E \left[ {x_{i,t}^2} \right] - 2{{\left( {\boldsymbol v_{ \cdot i}^{T}} \right)}^{new}} \mathbb E\left[ {{\boldsymbol y_t}} |\boldsymbol X_K \right] {x_{i,t}} \right. \nonumber\\
			& \left. + {{\left( {\boldsymbol v_{ \cdot i}^T} \right)}^{new}} \mathbb E\left[ {{\boldsymbol y_t} \boldsymbol y_t^T|\boldsymbol X_K} \right]{{\left( {{\boldsymbol v_{ \cdot i}}} \right)}^{new}} \right\} 
		\end{align}		
%	\end{equation}
	where $\left({\boldsymbol v_{ \cdot i}^T}\right)^{new}$ is the $i$th row of matrix $\boldsymbol V^{new}$, $ 1 \le i \le m $,  and $\boldsymbol \Sigma _x^{new} = diag\left( \left( {\sigma _1^2} \right)^{new},\cdots,\left( {\sigma _m^2} \right)^{new}\right)$.
	
	%	\quad
	
	With regard to $\boldsymbol \Sigma_1$, it is only contained in $P\left( {{\boldsymbol y_1}} \right)$, thus
	\begin{equation}\label{sigma_1}
		\begin{aligned}  % \limits_{P\left( {\boldsymbol Y_K|\boldsymbol X_K} \right)}
			{\boldsymbol \Sigma} _1^{new} 
			= &\arg \mathop {\max }\limits_{{\boldsymbol \Sigma _1}} \; \mathop \mathbb E \left[ \log {P\left( {{\boldsymbol y_1}|{\boldsymbol \Sigma _1}} \right)} \right]\\
			= &\mathbb E\left[ {{\boldsymbol y_1} \boldsymbol y_1^T|\boldsymbol X_K} \right]
		\end{aligned}
	\end{equation}
	
	For $\boldsymbol \Lambda= diag\left( {{\lambda _1}, \cdots ,{\lambda _p}} \right) $, $\boldsymbol \Sigma = \boldsymbol I-\boldsymbol \Lambda^2$.
	% to ensure the covariance matrix be the unit matrix. 
	 $\lambda_i$ is contained in $ {\Omega}_{\mathcal{M}_{K-1},i}^{\lambda} {{\left( {{\lambda _i} - {\lambda _{\mathcal{M}_{K-1},i}}} \right)}^2} $ and $  P\left( {{\boldsymbol y_t}|{\boldsymbol y_{t - 1}},\boldsymbol \Lambda } \right)  $, thus 
	\begin{equation}
	 {{\boldsymbol \Lambda ^{new}}} =  \arg \mathop {\max }\limits_ {\boldsymbol \Lambda} \;  J\left( \boldsymbol \Lambda\right) \nonumber
	\end{equation}
	where 
	%	\begin{equation}
	\begin{center}
		$	\begin{aligned}
			&J\left( \boldsymbol \Lambda\right)\\ 
			= &  \sum\limits_{t = 2}^{T_K} {\log \;P\left( {{\boldsymbol y_t}|{\boldsymbol y_{t - 1}},\boldsymbol \Lambda } \right)}  - \gamma_{2,K}  \sum\limits_{i = 1}^p {{\Omega}_{\mathcal{M}_{K-1},i}^{\lambda} {{\left( {{\lambda _i} - {\lambda _{\mathcal{M}_{K-1},i}}} \right)}^2}} \\
			=& - \frac{1}{2}\sum\limits_{t = 2}^{T_K} \sum\limits_{i = 1}^p \left[ \log \left( {1 - \lambda _i^2} \right) +  \frac{1}{{1 - \lambda _i^2}}\left( \mathbb E\left[ {y_{t,i}^2|\boldsymbol X_K} \right] \right. \right. \\
			&	\quad  \quad \quad \quad \;\; \;\;	\left.	\left.	- 2{\lambda _i} \mathbb E\left[ {{y_{t,i}}{y_{t - 1,i}}|\boldsymbol X_K} \right] + \lambda _i^2 \mathbb E\left[ {y_{t - 1,i}^2|\boldsymbol X_K} \right]  \right)  \right]  \\
			&  \quad  \quad \quad \quad \;\; \;\; \; - \gamma_{2,K}  \sum\limits_{i = 1}^p {{\Omega}_{\mathcal{M}_{K-1},i}^{\lambda} {{\left( {{\lambda _i} - {\lambda _{\mathcal{M}_{K-1},i}}} \right)}^2}}
		\end{aligned}$
	\end{center}

	Let the  derivative with respect $\lambda_i$ be zero, then
%	we drive the following equation
	\begin{equation}\label{lam_root}
		a_{i5} \lambda_i^5+a_{i4} \lambda_i^4+a_{i3} \lambda_i^3+ a_{i2} \lambda_i^2+a_{i1} \lambda_i+a_{i0}=0
	\end{equation}
	where 	the coefficients of \eqref{lam_root} are derived as 
	%	\begin{equation}
%	\left\{ {\begin{array}{*{20}{l}}
			\begin{align}
			a_{i5} = &2{\gamma _{2,K}}{{\Omega}_{\mathcal{M}_{K-1},i}^{\lambda}}, \nonumber \\
			a_{i4} = &- 2{\gamma _{2,K}}{{\Omega}_{\mathcal{M}_{K-1},i}^{\lambda}}{\lambda _{{\mathcal M}_{K-1},i}}, \nonumber \\ 
			a_{i3} = &{{T_K}-1 - 4{\gamma _{2,K}}{{\Omega}_{\mathcal{M}_{K-1},i}^{\lambda}}},\nonumber \\  
			a_{i2} = &{4{\gamma _{2,K}}{{\Omega}_{\mathcal{M}_{K-1},i}^{\lambda}}{\lambda _{{\mathcal M}_{K-1},i}} -\sum\limits_{t = 2}^{T_K}   \mathbb E\left[ {{y_{t,i}}{y_{t - 1,i}}|\boldsymbol X_K} \right]}, \nonumber \\ 
			a_{i1} = &{2{\gamma _{2,K}}{{\Omega}_{\mathcal{M}_{K-1},i}^{\lambda}} + \sum\limits_{t = 2}^{T_K} \left( \mathbb E\left[ {y_{t - 1,i}^2|\boldsymbol X_K} \right] + \mathbb E\left[ {y_{t,i}^2|\boldsymbol X_K} \right] - 1\right)},\nonumber\\  
			a_{i0} = & -{2{\gamma _{2,K}}{{\Omega}_{\mathcal{M}_{K-1},i}^{\lambda}}{\lambda_{\mathcal{M}_{K-1},i}} - \sum\limits_{t = 2}^{T_K} \mathbb E\left[ {{y_{t,i}}{y_{t-1,i}}|\boldsymbol X_K} \right]} \nonumber 
		\end{align} 
%\end{array}} \right. \nonumber	
%	\end{equation}
	Thus, the updated $\lambda_i^{new}$ could be calculated numerically as the root of \eqref{lam_root} within the range $[0,1)$, and $\boldsymbol \Lambda^{new}= diag\left( {{\lambda_1^{new}}, \cdots ,{\lambda_p^{new}}} \right) $.
	
	The learning procedure of PSFA--EWC is summarized in Algorithm \ref{PSFA-EWC_modeling}. The transformation  and emission matrices are denoted as $ {\boldsymbol \Lambda}_{\mathcal M_{K}} $ and $ {\boldsymbol V}_{\mathcal M_{K}} $, respectively.  Since noise information  about $ \boldsymbol \Sigma_1$ and $ \boldsymbol \Sigma_x$ is only effective for the current mode, the subscript $\mathcal{M}_K$ is neglected.  
	
	After the mode $\mathcal{M}_K$ has been learned, 
	%the importance measures should be updated by (\ref{importance_measure_V}--\ref{importance_measure_Lambda}). 	
	%Specifically when $K$th mode has been learned (see PSFA--EWC algorithm in Section \ref{sec3:2}), 
		the importance measures specific to PSFA  are updated  and ready as $(K+1)$th mode.
	\begin{equation}\label{importance_measure_V}
		{{\boldsymbol \Omega}_{\mathcal{M}_{K}}^V} = {{\boldsymbol \Omega}_{\mathcal{M}_{K-1}}^V} +  \eta_{K}^V { {\boldsymbol F}}^V_{\mathcal{M}_K}
	\end{equation}
	\begin{equation}\label{importance_measure_Lambda}
		{{\boldsymbol \Omega}_{\mathcal{M}_{K}}^{\Lambda}} = {{\boldsymbol \Omega}_{\mathcal{M}_{K-1}}^{\Lambda}} +  \eta_{{K}}^{\Lambda} { {\boldsymbol F}}^{\Lambda}_{\mathcal{M}_K}
	\end{equation}
	where $ { {\boldsymbol F}}^V_{\mathcal{M}_K} $ and $ { {\boldsymbol F}}^{\Lambda}_{\mathcal{M}_K} $ are  calculated by \eqref{fisher_V} and \eqref{fisher_lambda}.
	$ \eta_{K}^V $ and $ \eta_{K}^{\Lambda}$ are mode-specific hyperparamaters. 
	which are optimized by hyperparameter search  \cite{Kirkpatrick2017Overcoming} and fined-tune by prior knowledge,
	and  may play an  important role in accurate estimate of probability with sequential modes.  
		Then we illustrate the difference $ \gamma_{1,K} $ and $ \gamma_{2,K} $ to  $ \eta_{K-1}^{\Lambda} $ and $  \eta_{{K-1}}^{V} $. 	 Combined with  the importance of current mode $\mathcal{M}_K$, the setting $ \gamma_{1,K} $ and $ \gamma_{2,K} $ is beneficial to assigning the importance of all previous $K-1$ modes again. $ \eta_{K-1}^{\Lambda} $ and $  \eta_{{K-1}}^{V} $ focus on the importance of the mode $\mathcal{M}_{K-1}$, which allow users to obtain models with more focus on particular mode. 
		%Through the reasonable setting of four hyperparameters, the human-level performance may be obtained.

	\section{Monitoring procedure and experiment design}\label{sec:monitoring}
	Analogous to  traditional PSFA \cite{guo2016monitoring}, three monitoring statistics are designed to provide a comprehensive operating status. Then, several representative  methods  are adopted as  comparisons   to illustrate the superiorities of  PSFA--EWC algorithm.
	%\textcolor{red}{Add an introductory paragraph here}
	
		\begin{algorithm}[!bp]
		\caption{Off-line training procedure of PSFA--EWC}\label{PSFA-EWC_modeling}
	%	\small
			\footnotesize
		\begin{algorithmic}[1]
			\REQUIRE $\tilde{\boldsymbol X}_K$, $p$,  $ {\boldsymbol V}_{\mathcal M_{K-1}} $,   $ {\boldsymbol \Lambda}_{\mathcal M_{K-1}} $, $ {\boldsymbol \Omega}_{\mathcal M_{K-1}}^V $, ${\boldsymbol \Omega}_{\mathcal M_{K-1}}^ \Lambda $		
			\ENSURE  ${\boldsymbol \mu}_K$, $\boldsymbol \Sigma_K$, $ {\boldsymbol V}_{\mathcal M_{K}} $,   $ {\boldsymbol \Lambda}_{\mathcal M_{K}} $, $ {\boldsymbol \Omega}_{\mathcal M_{K}}^V $, ${\boldsymbol \Omega}_{\mathcal M_{K}}^ \Lambda $, $\boldsymbol K$, $\boldsymbol \Phi$, $ \boldsymbol \Xi $, $J_{th,T^2}$, $J_{th, SPE}$ and $J_{th,S^2}$
			% $ \boldsymbol \Sigma_1$, $ \boldsymbol \Sigma_x$, $ \boldsymbol \Sigma$				
			\STATE For the mode $\mathcal{M}_K$,  collect normal data  $\tilde{\boldsymbol X}_K$, calculate the mean ${\boldsymbol \mu}_K$ and standard variance $\boldsymbol \Sigma_K$. Normalize data and  the scaled data are denoted as $\boldsymbol X_K$;
			\STATE Initialize parameters $ \boldsymbol V$,  $ \boldsymbol \Lambda$, $ \boldsymbol \Sigma_1$, $ \boldsymbol \Sigma_x$, $ \boldsymbol \Sigma$;			
			\STATE  \textbf{While} the issue \eqref{PSFAWC_aim} is not converged \textbf{do}
			\begin{enumerate}[a)]				
				\item Calculate three sufficient statistics by \eqref{mean}--\eqref{variance};
				% based on Algorithm \ref{alg:estep} in Appendix B; %\ref{alg:estep};
				
				\item Update the parameters by  \eqref{first_dev_v}--\eqref{lam_root};%(\ref{first_dev_v}, \ref{sigma_x}, \ref{sigma_1}, \ref{lam_root});
			\end{enumerate}
			
			\STATE  The optimal emission and transition  matrices are denoted as  $ {\boldsymbol V}_{\mathcal M_{K}} $ and $ {\boldsymbol \Lambda}_{\mathcal M_{K}} $, respectively;				
			
			\STATE Calculate the FIMs ${{\boldsymbol F}}_{\mathcal M_{K}}^V $ by \eqref{fisher_V} and ${{\boldsymbol F}}_{\mathcal M_{K}}^\Lambda $ by  \eqref{fisher_lambda}. Then, update the importance measures   $ {\boldsymbol \Omega}_{\mathcal M_{K}}^V $ and  ${\boldsymbol \Omega}_{\mathcal M_{K}}^ \Lambda $  by  \eqref{importance_measure_V}--\eqref{importance_measure_Lambda};
			
			\STATE The final Kalman matrix  is denoted as $\boldsymbol K$,		calculate $\boldsymbol \Phi$ and $\boldsymbol \Xi$;	
			
			\STATE Calculate three monitoring statistics by  \eqref{T2}, \eqref{SPE}, \eqref{S2};
			
			\STATE  Calculate  thresholds by KDE, labeled as $J_{th,T^2}$, $J_{th, SPE}$ and $J_{th,S^2}$.
			
		\end{algorithmic}
	\end{algorithm} 

	\subsection{Monitoring procedure}	
	
	%Analogous to PSFA, three monitoring statistics are designed for PSFA-EWC, where 
	In this paper, the Hotelling's $T^2$ and $ SPE $ statistics are used to reflect the steady variations, and $S^2$ is calculated to evaluate the temporal dynamics\cite{guo2016monitoring}.
	% and distinguish the normal variations and the real faults 
	
	According to Kalman filter equation,
	\begin{equation}\label{latentvariabley}
		\boldsymbol	y_t = {\boldsymbol \Lambda}_{\mathcal{M}_{K}} \boldsymbol y_{t-1} + \boldsymbol K\left[\boldsymbol x_t - {\boldsymbol V}_{\mathcal{M}_{K}} {\boldsymbol \Lambda}_{\mathcal{M}_{K}} \boldsymbol y_{t-1}\right]
	\end{equation}
After the training phase,  $\boldsymbol K_t$ would converge to a steady matrix $\boldsymbol K$
%	When $t \to \infty $,  $\boldsymbol K_t$ converges to a steady matrix $\boldsymbol K$. $\boldsymbol K_t $ is stable after the training phase. 
 Then, $T^2$ statistic is defined as
	\begin{equation}\label{T2}
		T^2 = \boldsymbol y_t^T \boldsymbol y_t
	\end{equation} 
	
	To design $SPE$,  the bias between the true value and one-step prediction is calculated at $t$ instant. At $t-1$ instant, the inferred slow features follow Gaussian distribution, namely,
	\begin{equation}
		P\left( \boldsymbol y_{t-1}|\boldsymbol x_{1},\cdots,\boldsymbol x_{t-1} \right)  \sim  N \left(\boldsymbol \mu_{t-1}, \boldsymbol P_{t-1}\right) \nonumber
	\end{equation}
	Then, the conditional distribution of $\boldsymbol y_{t}$ is described as
	%\begin{center}
	\begin{equation}
		P\left( \boldsymbol y_{t}|\boldsymbol x_{1},\cdots,\boldsymbol x_{t-1} \right)  \sim  N \left( {\boldsymbol \Lambda}_{\mathcal{M}_{K}} \boldsymbol \mu_{t-1}, {\boldsymbol \Lambda}_{\mathcal{M}_{K}} \boldsymbol P_{t-1} {\boldsymbol \Lambda}_{\mathcal{M}_{K}}^T+ \boldsymbol \Sigma \right) \nonumber
	\end{equation}
	%	\end{center}
	Similarly, 
	%	$	P\left( \boldsymbol x_{t}|\boldsymbol x_{1},\cdots,\boldsymbol x_{t-1} \right)  \sim  N \left( {\boldsymbol  V}_{\mathcal{M}_{K}}{\boldsymbol \Lambda}_{\mathcal{M}_{K}} \boldsymbol \mu_{t-1}, \boldsymbol \Phi_t \right)$, 
	\begin{center}
		$	P\left( \boldsymbol x_{t}|\boldsymbol x_{1},\cdots,\boldsymbol x_{t-1} \right)  \sim  N \left( {\boldsymbol  V}_{\mathcal{M}_{K}}{\boldsymbol \Lambda}_{\mathcal{M}_{K}} \boldsymbol \mu_{t-1}, \boldsymbol \Phi_t \right)$
	\end{center}
	where	$ \boldsymbol \Phi_t = \boldsymbol V_{\mathcal{M}_{K}} {\boldsymbol \Lambda}_{\mathcal{M}_{K}} \boldsymbol P_{t-1} {\boldsymbol \Lambda}_{\mathcal{M}_{K}}^T \boldsymbol V_{\mathcal{M}_{K}}^T +\boldsymbol V_{\mathcal{M}_{K}}  \boldsymbol \Sigma \boldsymbol V_{\mathcal{M}_{K}}^T+ \boldsymbol \Sigma_x $. The prediction error follows Gaussian distribution, namely
	\begin{equation}\label{prediction_error}
		\boldsymbol \varepsilon_t = \boldsymbol x_t -   {\boldsymbol  V}_{\mathcal{M}_{K}} {\boldsymbol \Lambda}_{\mathcal{M}_{K}} \boldsymbol \mu_{t-1} \sim N \left( \boldsymbol  0, \boldsymbol \Phi_t \right)
	\end{equation}
	%When $t \to \infty $,  
	After the training phase, $ \boldsymbol \Phi_t  $ converges to $ \boldsymbol \Phi $. The $ SPE $ statistic is calculated by
	\begin{equation}\label{SPE}
		SPE = \boldsymbol \varepsilon_t^T \boldsymbol \Phi^{-1} \boldsymbol \varepsilon_t
	\end{equation}
	
	$S^2$ statistic is designed to reflect the temporal dynamics, which is beneficial to distinguishing the normal operating variations and dynamics anomalies \cite{shang2015concurrent,guo2016monitoring}. 
	\begin{equation}\label{S2}
		S^2 = \dot {\boldsymbol y}_t^T{\boldsymbol \Xi ^{ - 1}}{{\dot {\boldsymbol y}}_t}
	\end{equation}
	where  $ \dot {\boldsymbol y}_t = {\boldsymbol y}_t - {\boldsymbol y}_{t-1} $, $\boldsymbol \Xi  = {\mathbb E}\left\{ {{{\dot {\boldsymbol  y}}_t}\dot {\boldsymbol y}_t^T} \right\}$ is the covariance matrix and analytically calculated as $	{\boldsymbol{\Xi }} = 2\left( {{\boldsymbol I_p} - {\boldsymbol \Lambda}_{\mathcal{M}_{K}} } \right)$ \cite{guo2016monitoring}.

		\begin{algorithm}[!tbp]
		\caption{Online monitoring procedure of PSFA--EWC}\label{PSFA-EWC_monitoring}
		\footnotesize
%	\small
		\begin{algorithmic}[1]
			\STATE Collect the test data $\boldsymbol x$, preprocess  $\boldsymbol x$ by ${\boldsymbol \mu}_K$ and  $\boldsymbol \Sigma_K$;
			\STATE Calculate the latent variable by \eqref{latentvariabley} and prediction error by \eqref{prediction_error};
			\STATE Calculate three monitoring statistics by \eqref{T2}, \eqref{SPE}, \eqref{S2};
			\STATE Judge the operating condition: 
			\begin{enumerate}[a)]
				\item Normal, return to step 1; 
				\item A new mode appears, let $K=K+1$, return to Algorithm \ref{PSFA-EWC_modeling} to update the monitoring model; 
				\item A fault occurs and the alarm is triggered.
			\end{enumerate}			
		\end{algorithmic}
	\end{algorithm} 

	The thresholds of three statistics are calculated by kernel density estimation (KDE) \cite{zhang2019an}, and denoted as $J_{th,T^2}$, $J_{th,SPE}$ and $J_{th,S^2}$. The monitoring rule is summarized below:
	\begin{enumerate}
		\item All statistics are within thresholds, the process is normal; 
		
		\item If $T^2$ or $ SPE $ is over its threshold, while $S^2$ is below its threshold,  the dynamic law remains unchanged and the static variations occur. This may be caused by step faults or normal drifts  \cite{shang2015concurrent,yu2019recursive}, which should be confirmed further based on data and limited prior knowledge. When a new mode occurs, a small set of new data are collected to update the PSFA--EWC model.  The process is monitored by $S^2$ statistic before the updating procedure;
		
		\item If  $S^2$ is over threshold, the dynamic behaviors are unusual and the  system is out of control. A fault occurs and the alarm would be triggered.
		
	\end{enumerate}
	
	The off-line training and online monitoring procedures  have been summarized in Algorithm \ref{PSFA-EWC_modeling} and Algorithm \ref{PSFA-EWC_monitoring}, respectively. Fault detection rates (FDRs) and false alarm rates (FARs) are adopted to evaluate the performance.
	% When a new mode is identified, assume that a set of data are collected 

	\subsection{Comparative design}
	
	%To illustrate the effectiveness and superiorities of the proposed PSFA-EWC, 
	%In this paper,
	 RSFA \cite{shang2018recursive},
	PCA--EWC \cite{zhang2021multimode}, IMPPCA \cite{zhang2019an} and MCVA \cite{wen2015multimode} are selected as the comparative methods in Table \ref{comparative-design}.  
    For PSFA--EWC, PSFA and RSFA, three monitoring statistics are calculated, where $S^2$
	 statistic is beneficial to distinguishing real faults and normal dynamic behaviors in multimode nonstationary processes.
	% is established to reflect the dynamic behaviors and the occurrence of a fault is confirmed for multimode processes.
	The remaining methods calculate two statistics and cannot separate real faults from normal variations. Assume that data from each mode are collected sequentially, the performance is evaluated by monitoring the current and the previously learned modes. 
%	a fault is detected when $ SPE $ or $T^2$ is beyond the corresponding threshold.

	\begin{table}[!tbp]
		\begin{center}
			\caption{Comparative schemes}\label{comparative-design}
			\footnotesize
		%	\renewcommand\arraystretch{1}
			%   \small
			\begin{tabular}{c c c c c}
				\hline
				& Methods &  \makecell{Training sources\\(Model + Data)}  & \makecell{Model\\ label} & \makecell{Testing \\sources}  \\
				\hline
				Situation 1 & PSFA       &$\mathcal{M}_1$            &  A     & $\mathcal{M}_1$      \\
				Situation 2 & PSFA--EWC   & A + $\mathcal{M}_2$        &  B     & $\mathcal{M}_2$     \\
				Situation 3 & PSFA--EWC    &  -                         & B     &  $\mathcal{M}_1$   \\
				Situation 4 & PSFA      &  $\mathcal{M}_2$           & C     &  $\mathcal{M}_2$     \\
				Situation 5 & PSFA       &  -                & C       &$\mathcal{M}_1$      \\
				Situation 6 & PSFA--EWC    &  B + $\mathcal{M}_3$   &  D       &$\mathcal{M}_3$    \\
				Situation 7 & PSFA--EWC    &  -                &  D       & $\mathcal{M}_1$    \\
				Situation 8 & PSFA--EWC   &  -                & D       &  $\mathcal{M}_2$    \\
				Situation 9 & PSFA       &   $\mathcal{M}_3$           & E       & $\mathcal{M}_3$    \\
				Situation 10 & PSFA     & -                 &E       & $\mathcal{M}_1$      \\
				Situation 11 & PSFA     & -                 & E       & $\mathcal{M}_2$      \\			
				\hline
				Situation 12& RSFA     & $\mathcal{M}_1$          &  F     & $\mathcal{M}_1$    \\
				Situation 13& RSFA     &  F + $\mathcal{M}_2$     &  G     &$\mathcal{M}_2$    \\
				Situation 14& RSFA     &  G + $\mathcal{M}_3$      &  H     & $\mathcal{M}_3$     \\
				\hline
				Situation 15&  PCA        & $\mathcal{M}_1$          &  I     & $\mathcal{M}_1$    \\
				Situation 16& PCA--EWC     &  I + $\mathcal{M}_2$     &  J     &$\mathcal{M}_2$    \\
				Situation 17& PCA--EWC     &  -                  &  J     & $\mathcal{M}_1$     \\
				Situation 18& PCA--EWC     & J + $\mathcal{M}_3$   &  L     & $\mathcal{M}_3$     \\
				Situation 19& PCA--EWC     & -                   &  L     & $\mathcal{M}_1$     \\
				Situation 20& PCA--EWC     & -                   &  L     & $\mathcal{M}_2$    \\
				\hline
				Situation 21 & IMPPCA   & $\mathcal{M}_1$, $\mathcal{M}_2$      & M      & $\mathcal{M}_1$    \\
				Situation 22 & IMPPCA   &  -                    &  M      & $\mathcal{M}_2$    \\
				Situation 23 & IMPPCA   & $\mathcal{M}_1$, $\mathcal{M}_2$, $\mathcal{M}_3$   &  N    & $\mathcal{M}_1$ \\
				Situation 24 & IMPPCA   &  -                    &  N      & $\mathcal{M}_2$     \\
				Situation 25 & IMPPCA   &  -                    & N      & $\mathcal{M}_3$     \\
				\hline
				Situation 26 & MCVA     & $\mathcal{M}_1$, $\mathcal{M}_2$    &  O      & $\mathcal{M}_1$      \\
				Situation 27 & MCVA     &  -                     & O      & $\mathcal{M}_2$     \\
				Situation 28 & MCVA     & $\mathcal{M}_1$, $\mathcal{M}_2$, $\mathcal{M}_3$    & P     & $\mathcal{M}_1$ \\
				Situation 29 & MCVA     &  -                     & P      & $\mathcal{M}_2$     \\
				Situation 30 & MCVA     &  -                     &  P      & $\mathcal{M}_3$      \\
				\hline
			\end{tabular}
		\end{center}
	\end{table}
	
	PSFA--EWC, RSFA and PCA--EWC can be regarded as adaptive methods, which avoid storing data and alleviating storage requirement.
	For Situations 1--11, PSFA and PSFA--EWC are compared to illustrate the catastrophic forgetting issue of PSFA and the continual learning ability of PSFA--EWC for successive nonstationary modes. When a new mode  is identified by $S^2$ statistic and limited prior knowledge, a small set of normal data are collected and the model is updated off-line by extracting new information while consolidating the learned knowledge. %The extracted information is sufficient to monitor multiple modes simultaneously.
	 PSFA--EWC furnishes the backward and forward transfer ability. The updated model is able to monitor the previous modes and the learned knowledge is valuable to learn new relevant modes.
	%Since the significant information from modes has been extracted, the model enables to monitors two modes successfully. 
	%	Similarly, when a new mode is detected, the model is learned based on the aforementioned scheme and can monitor more modes simultaneously. 
	Equivalently, the simulation results of Situations 2, 3, 6--8  should be excellent. Conversely, the results of Situations 5, 10 and 11 are expected to be poor.
	% thus the catastrophic forgetting issue is reflected.  
    %RSFA is selected to monitor the sequential models. 
	The RSFA  model is updated in real time and desired to track the system adaptively, as Situations 12--14 illustrated. For Situations 15--20, the design process of PCA--EWC is similar to that of PFSA--EWC. PCA--EWC is desired to provide the  continual learning ability comparable to PSFA--EWC.
	
	 IMPPCA and MCVA are multiple-model methods, where the mode is identified and local models are built within each mode. 
	Data from all possible  modes are required before learning. When a novel mode arrives,  sufficient samples should be collected and the model needs to be retrained on the entire dataset. 
	%For example, when mode $\mathcal{M}_3$ appears, the model is relearned based on data from three modes. 
	%The model can deliver optimal monitoring consequences for the learned modes. Thus, 
	IMPPCA and MCVA should provide excellent performance for Situations 21--30.  
	%Since data from all modes need to be stored, 
	 However, it is intractable and time-consuming to collect complete data in practical systems \cite{quinones-grueiro2019data}.
	The computational resources would increase for each retraining with the increasing number of modes. 
	%Since IMPPCA and MCVA share the similar procedure, take IMPPCA as an example to illustrate. When two modes are available, the model is learned based on all normal data. When a new mode appears, the model is retrained based on data from three modes.
	%The model parameters in each mode are determined when the optimization issue is settled.

	\begin{table}[!hbp]	
	\begin{center}
		\caption{Normal operating modes and data information of CSTH}\label{Data_informationCSTH}
		\footnotesize
		\renewcommand\arraystretch{1}
		%		\small
		\begin{tabular}{c c c c | c c } %{c c c c}
			\hline
			&\multicolumn{3}{c|} {\makecell{Normal operating setting}} &   \multicolumn{2}{c} {\makecell{Data information}} \\
			\hline
			\makecell{Mode\\ number} & \makecell{Level\\ SP} &\makecell{Temperature \\ SP} & \makecell{Hot water\\ valve} & \makecell{NoTrS}  & \makecell{NoTeS}\\
			\hline
			\makecell{ ${\mathcal M}_1$}  & 9  & 10.5  & 4.5 & 1000  &1000\\
			\makecell{ ${\mathcal M}_2$} &12   & 8    & 4   & 300  &1000\\
			\makecell{${\mathcal M}_3$}  &12  &10.5    & 5.5 & 300   & 1000\\ 							
			\hline	
		\end{tabular}
	\end{center}
\end{table}

\begin{table*}[!htbp]
	\begin{center}
		\caption{FDRs ($\%$) and FARs ($\%$) for PSFA, PSFA--EWC and RSFA}\label{Table-PSFAEWC}
		\footnotesize
	%\small
		%	\renewcommand\arraystretch{1.3}
		%	\setlength\tabcolsep{3.5pt}%调列距
		\begin{tabular}{l c c c c c c c c c c c c c c  c c c c c}
			\hline
			\multirow{3}*{\makecell{Fault\\ type}} 	&\multirow{3}*{Method}   
			&\multicolumn{6}{c}{CSTH}   &\multicolumn{6}{c}{Pulverizing system} \\
			\cmidrule(r){3-8} \cmidrule(r){9-14}% \cmidrule(r){15-20}
			&	& \multicolumn{2}{c}{$T^2$}  & \multicolumn{2}{c}{$ SPE $}  & \multicolumn{2}{c} {$S^2$} & \multicolumn{2}{c}{$T^2$}  & \multicolumn{2}{c}{$ SPE $}  & \multicolumn{2}{c} {$S^2$} \\ % & \multicolumn{2}{c}{$T^2$}  & \multicolumn{2}{c}{SPE}  & \multicolumn{2}{c} {$S^2$} \\	
			\cmidrule(r){3-4}  \cmidrule(r){5-6}  \cmidrule(r){7-8}  \cmidrule(r){9-10}  \cmidrule(r){11-12}  \cmidrule(r){13-14} 
			% \cmidrule(r){15-16}  \cmidrule(r){17-18}  \cmidrule(r){19-20} 
			&	& FDR &FAR    &FDR &FAR     &FDR  &FAR      & FDR &FAR    & FDR &FAR     &FDR &FAR    \\
			%&FDR & FAR      & FDR & FAR     & FDR & FAR \\		    
			\hline			  
			Situation 1  &PSFA       &43.6 &7.4     &95.0 &1.4    &77.0 &0.2     &99.92  &2.15   &99.92  &0      &20.05  &0.86\\
			Situation 2  &PSFA--EWC  &91.2 &0       &89.6 &0      &98.0 &0.2     &100    &9.74   &100   &4.07    &94.71  &3.59\\
			Situation 3  &PSFA--EWC  &92.0 &0       &92.4 &0      &99.2 &9.8     &99.92  &1.29   &99.92  &1.29   &94.01  &14.04\\
			Situation 4  &PSFA       &1.2 & 0       &93.4 &0.6    &0.2 &0.2      &100    &48.80  &100  &5.91    &93.25  &0.40\\
			Situation 5  &PSFA       &4.2 &6.2      &95.0 &0.8    &20.0 &21.8    &100    &15.19  &99.92  &0     &93.13  &14.04 \\
			Situation 6  &PSFA--EWC  &88.6 &0       &89.4 &1.2    &98.4 &1.6     &100   &3.65    &100   &0.20    &94.73  &3.55\\
			Situation 7  &PSFA--EWC  &90.6 &0       &93.4 &0.8    &99.0 &0.8     &99.92  &1.00   &99.92  &0      &88.10  &10.89\\
			Situation 8  &PSFA--EWC  &89.8 &        &92.0 &5.8    &98.2 &0.2     &100   &7.91    &99.45  &0.24   &95.26  &0.88\\
			Situation 9  &PSFA       &68.8 &23.2    &94.4 &8.0    &78.8 &0.2     &100   &1.32    &100  &0.20     &53.63  &2.13 \\
			Situation 10 &PSFA       &54.0 &3.6     &95.4 &5.6    &83.4 &0.4     &99.92  &66.05  &99.92  &0.29   &83.95  &9.60\\
			Situation 11 &PSFA       &100 &100      &96.4 &74.4   &83.2 &2.4     &100   &79.63   &99.45  &0     &91.79   &0.24\\
			\hline 
			Situation 12 &RSFA      &0   &0     &0   &0      &42.0   &0.6        &60.30   &0      & 0 &0     &43.13   & 1.86 \\ 
			Situation 13 &RSFA      &0    &0     &0   &0      &59.2   &1.0         &84.49   &0      &0  &0     &37.59   &1.68  \\  
			Situation 14 &RSFA      &0   &0     &0   &0      &28.8   &0.2         & 0  &0         &1.98  &0  &6.59   &1.02  \\    
			\hline
		\end{tabular}
	\end{center}
\end{table*}

\section{Case studies}\label{sec4}
\subsection{CSTH case}	
% xu2014multimode,
The CSTH process is a nonlinear nonstationary process and  widely utilized as a benchmark for multimode process monitoring  \cite{quinones-grueiro2019data,Huang2020structure}.  Thornhill \textit{et al.} built the CSTH model and the detail information was described in  \cite{thornhill2008347A}. %The CSTH  process  mixes the hot water and cold water well to satisfy the demand.
CSTH aims to mix the hot and cold water with desirable settings. 
Level, temperature and flow are manipulated by PI controllers.   Six critical variables are selected for monitoring  and three successive modes are considered  in Table \ref{Data_informationCSTH}. 
%For convenience,  successive
The numbers of training and testing samples are denoted as NoTrS and NoTeS, respectively. A random fault occurs in the level from the 501$ th $ sample and the fault amplitude is 0.15.

\begin{figure*}[!tbp]
	\centering
	\subfigure{\label{fault3-2}}\addtocounter{subfigure}{-2}
	\subfigure
	{\subfigure[Situation 2]{\includegraphics[width=0.236\textwidth]{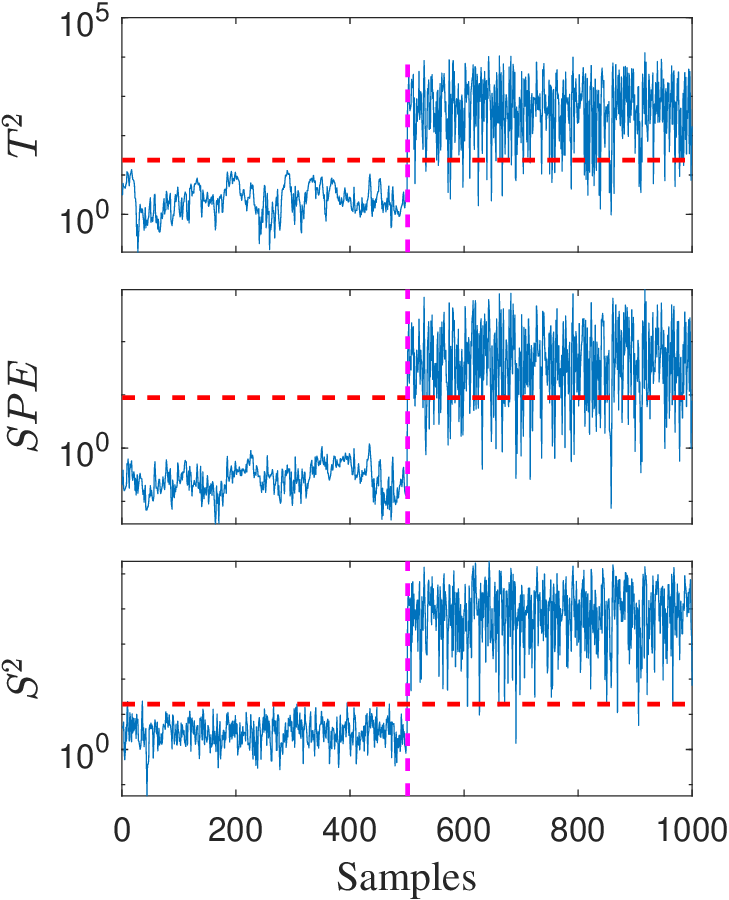}}}
	%	\vspace{-0.5mm}
	%	\hspace{-1mm}
	%	\subfigure{\label{fault3-3}}\addtocounter{subfigure}{-2}
	%	\subfigure
	%	{\subfigure[Situation 3]{\includegraphics[width=0.236\textwidth]{PSFAEWC_mode12_mode1_Fault1.eps}}}
	\vspace{-0.5mm}
	\hspace{-1mm}
	\subfigure{\label{fault3-4}}\addtocounter{subfigure}{-2}
	\subfigure
	{\subfigure[Situation 4]{\includegraphics[width=0.236\textwidth]{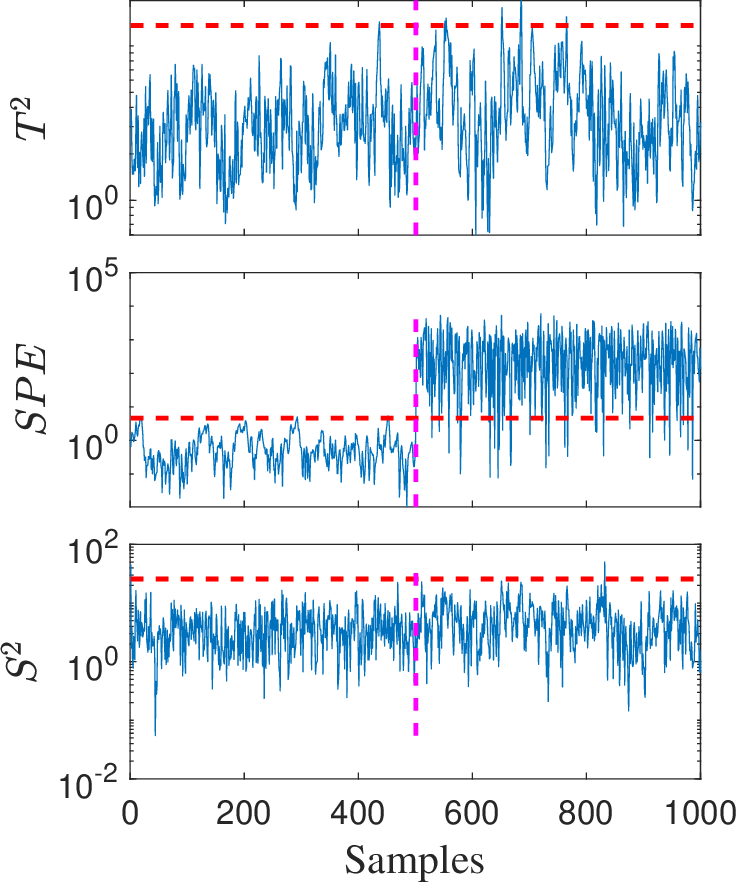}}}
	%	\vspace{-0.5mm}
	%	\hspace{-1mm}
	%	\subfigure{\label{fault3-5}}\addtocounter{subfigure}{-2}
	%	\subfigure
	%	{\subfigure[Situation 5]{\includegraphics[width=0.236\textwidth]{PSFAEWC_mode2_mode1_Fault1.eps}}}
	%
	\vspace{-0.5mm}
	\hspace{-1mm}
	\subfigure{\label{fault3-6}}\addtocounter{subfigure}{-2}
	\subfigure
	{\subfigure[Situation 6]{\includegraphics[width=0.236\textwidth]{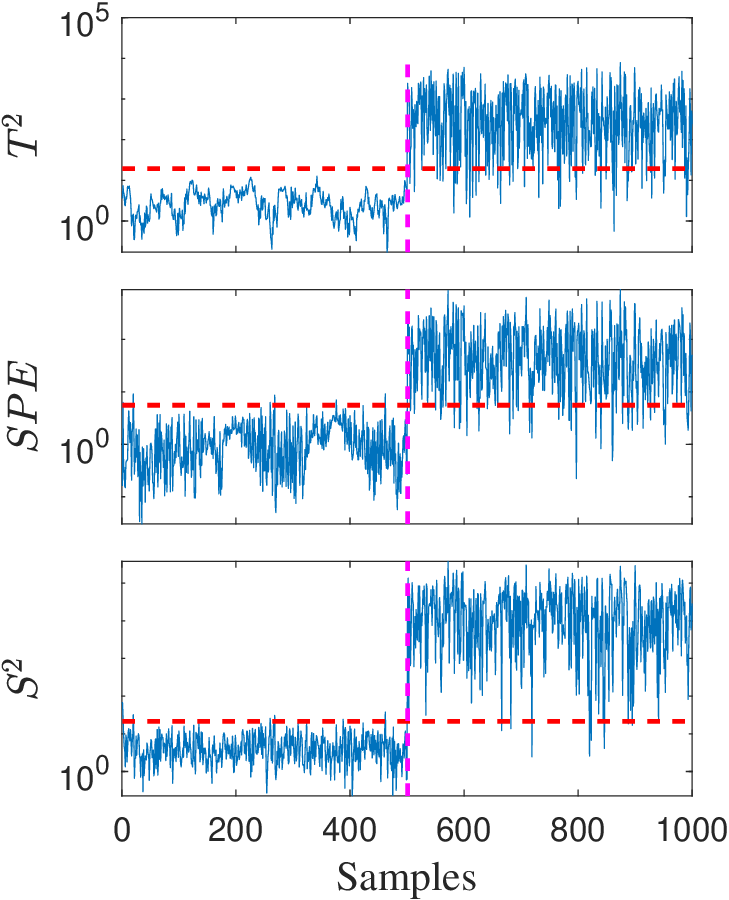}}}
	\vspace{-0.5mm}
	\hspace{-1mm}
	\subfigure{\label{fault3-7}}\addtocounter{subfigure}{-2}
	\subfigure
	{\subfigure[Situation 7]{\includegraphics[width=0.236\textwidth]{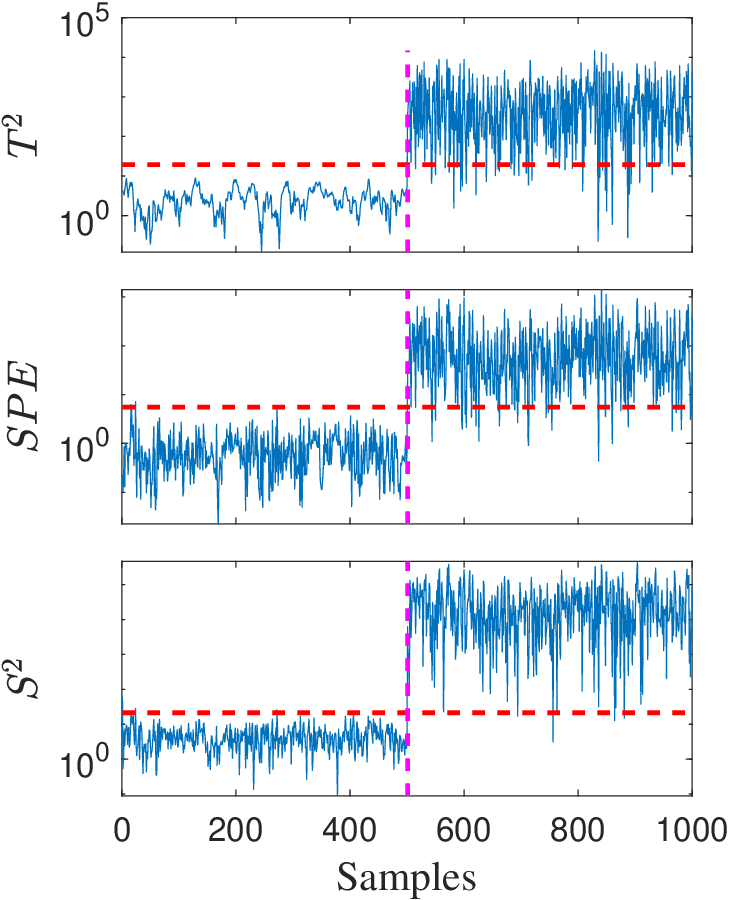}}}
	\vspace{-0.5mm}
	\hspace{-1mm}   	
	\subfigure{\label{fault3-8}}\addtocounter{subfigure}{-2}
	\subfigure
	{\subfigure[Situation 8]{\includegraphics[width=0.236\textwidth]{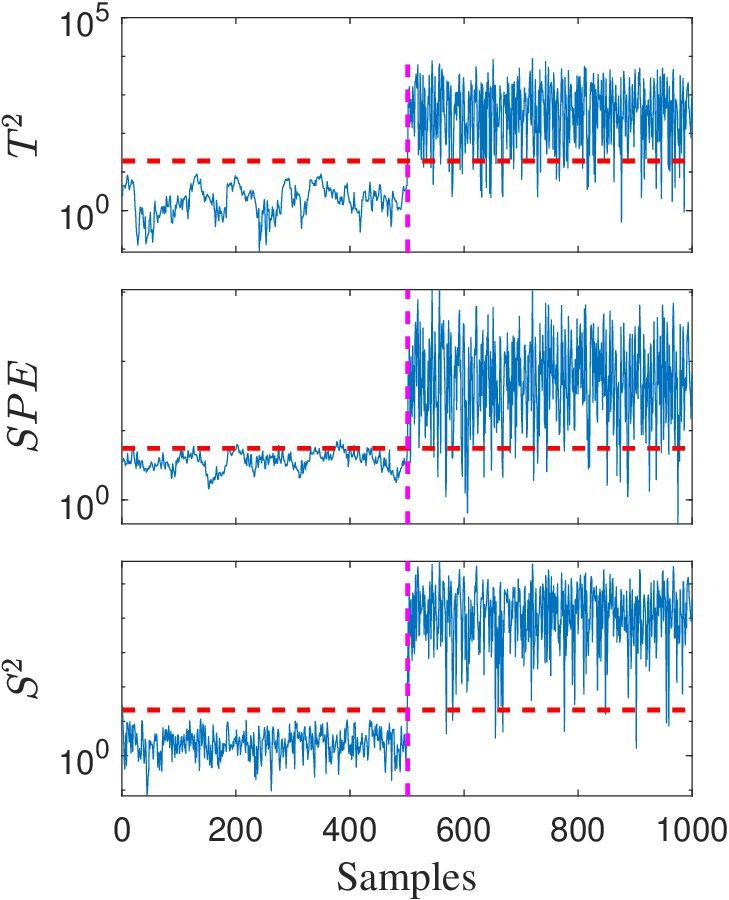}}}
	\vspace{-0.5mm}
	\hspace{-1mm}   	
	\subfigure{\label{fault3-9}}\addtocounter{subfigure}{-2}
	\subfigure
	{\subfigure[Situation 9]{\includegraphics[width=0.236\textwidth]{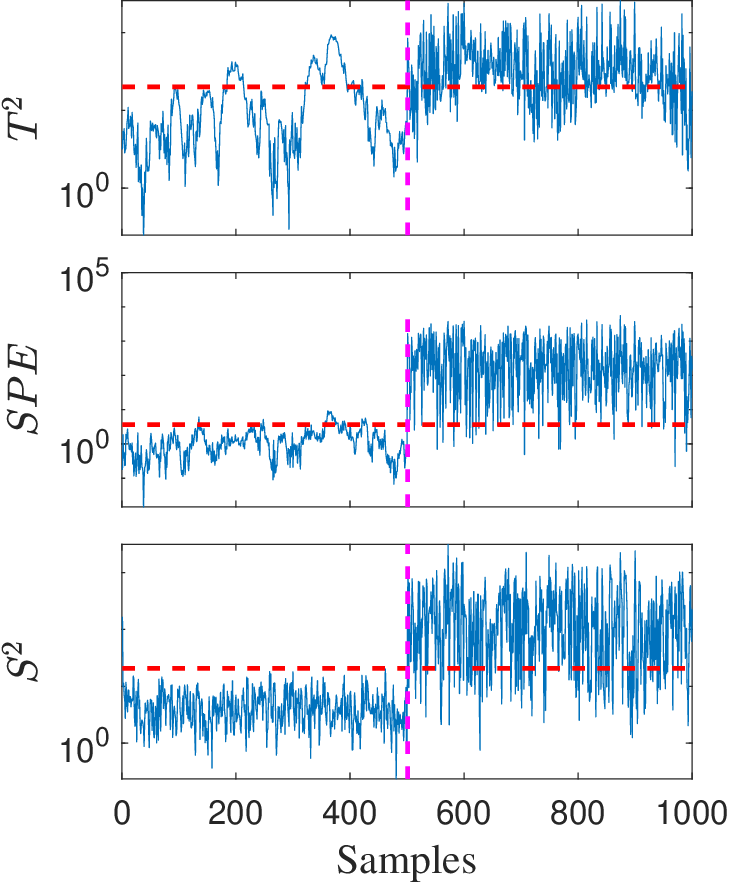}}}
	%	\vspace{-0.5mm}
	%	\hspace{-1mm}
	%	\subfigure{\label{fault3-10}}\addtocounter{subfigure}{-2}
	%	\subfigure
	%	{\subfigure[Situation 10]{\includegraphics[width=0.236\textwidth]{PSFAEWC_mode3_mode1_Fault1.eps}}}
	\vspace{-0.5mm}
	\hspace{-1mm}
	\subfigure{\label{fault3-11}}\addtocounter{subfigure}{-2}
	\subfigure
	{\subfigure[Situation 11]{\includegraphics[width=0.236\textwidth]{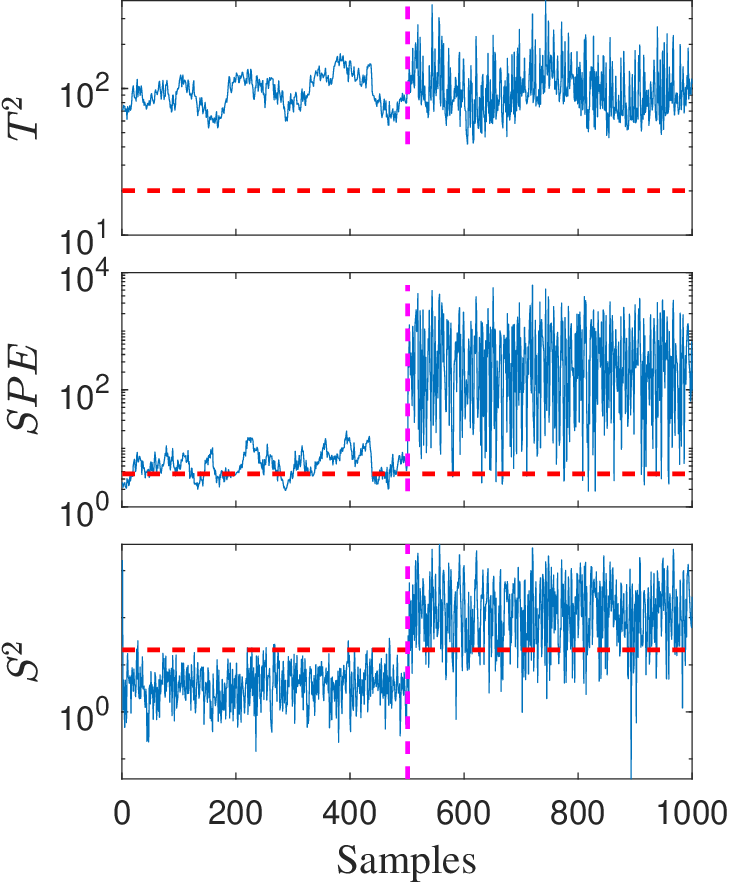}}}
	%	\vspace{-0.5mm}
	%	\hspace{-1mm}
	%	\subfigure{\label{fault3-13}}\addtocounter{subfigure}{-2}
	%	\subfigure
	%	{\subfigure[Situation 13]{\includegraphics[width=0.236\textwidth]{RSFA_mode12_mode2_Fault1.eps}}}
	\vspace{-0.5mm}
	\hspace{-1mm}
	\subfigure{\label{fault3-14}}\addtocounter{subfigure}{-2}
	\subfigure
	{\subfigure[Situation 14]{\includegraphics[width=0.236\textwidth]{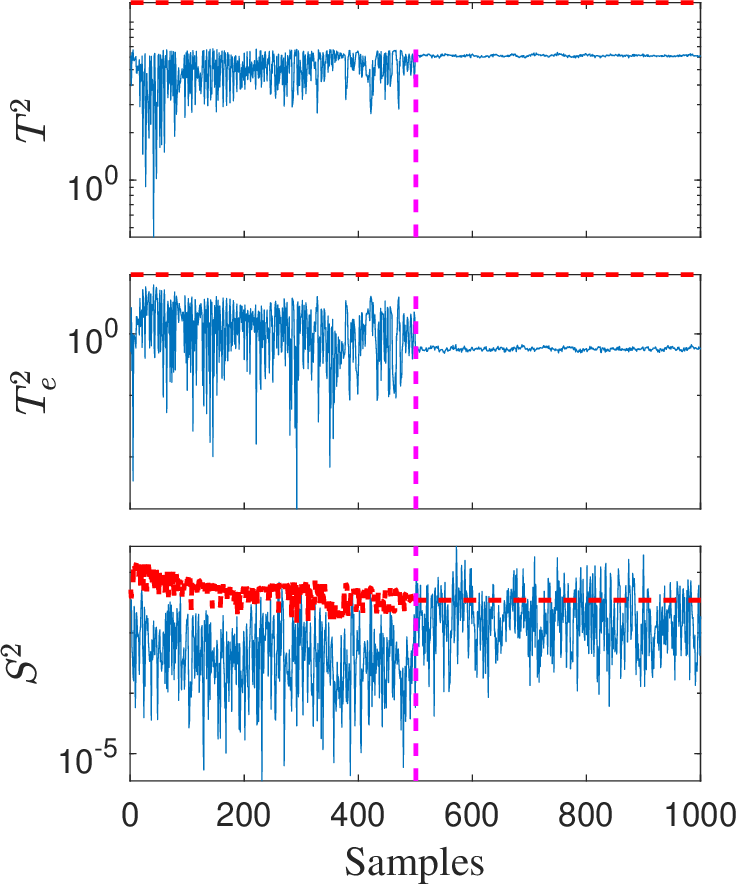}}}
	%
	%	\vspace{-0.5mm}
	%	\hspace{-1mm}
	%	\subfigure{\label{fault3-16}}\addtocounter{subfigure}{-2}
	%	\subfigure
	%	{\subfigure[Situation 16]{\includegraphics[width=0.236\textwidth]{PCAEWC_mode12_mode2_Fault1.eps}}}
	%	%
	%	\vspace{-0.5mm}
	%	\hspace{-1mm}
	%	\subfigure{\label{fault3-17}}\addtocounter{subfigure}{-2}
	%	\subfigure
	%	{\subfigure[Situation 17]{\includegraphics[width=0.236\textwidth]{PCAEWC_mode12_mode1_Fault1.eps}}}
	%
	\vspace{-0.5mm}
	\hspace{-1mm}
	\subfigure{\label{fault3-18}}\addtocounter{subfigure}{-2}
	\subfigure
	{\subfigure[Situation 18]{\includegraphics[width=0.236\textwidth]{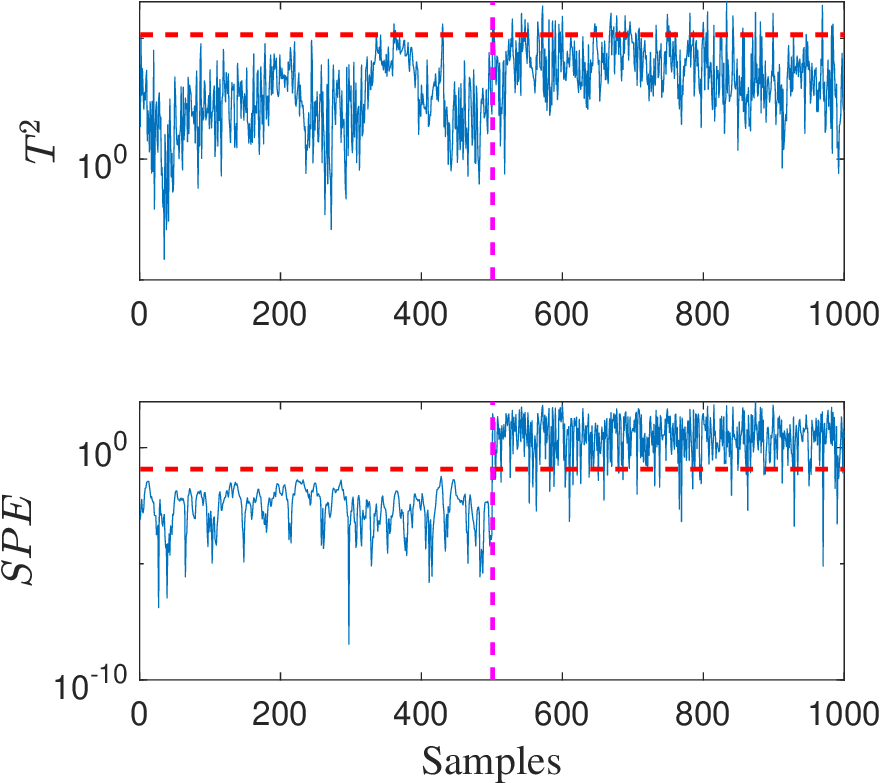}}}
	%
	%	\vspace{-0.5mm}
	%	\hspace{-1mm}
	%	\subfigure{\label{fault3-19}}\addtocounter{subfigure}{-2}
	%	\subfigure
	%	{\subfigure[Situation 19]{\includegraphics[width=0.236\textwidth]{PCAEWC_mode123_mode1_Fault1.eps}}}
	%
	\vspace{-0.5mm}
	\hspace{-1mm}
	\subfigure{\label{fault3-20}}\addtocounter{subfigure}{-2}
	\subfigure
	{\subfigure[Situation 20]{\includegraphics[width=0.236\textwidth]{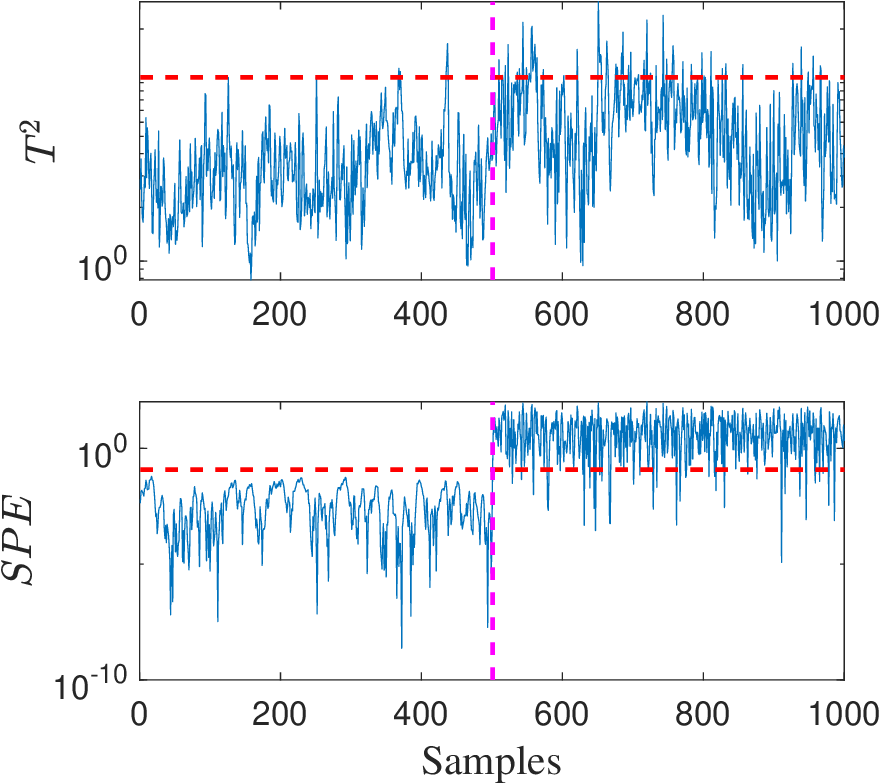}}}
	%
	%	\vspace{-0.5mm}
	%	\hspace{-1mm}
	%	\subfigure{\label{fault3-22}}\addtocounter{subfigure}{-2}
	%	\subfigure
	%	{\subfigure[Situation 22]{\includegraphics[width=0.236\textwidth]{IMPPCA_mode12_mode2_Fault1fig2.eps}}}
	%
	\vspace{-0.5mm}
	\hspace{-1mm}
	\subfigure{\label{fault3-23}}\addtocounter{subfigure}{-2}
	\subfigure
	{\subfigure[Situation 23]{\includegraphics[width=0.236\textwidth]{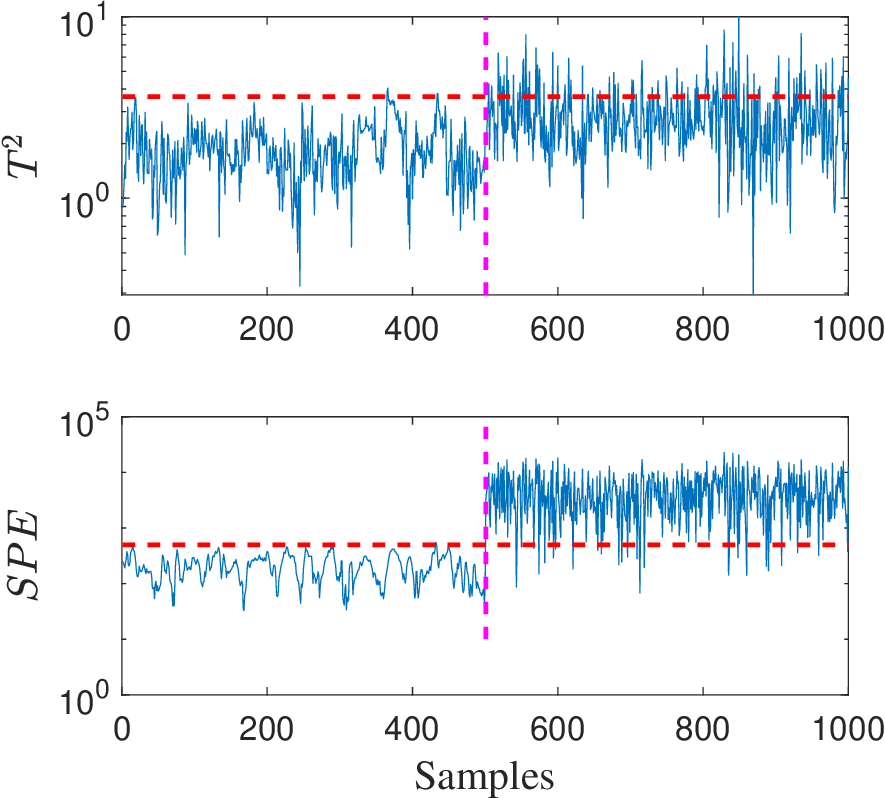}}}
	\vspace{-0.5mm}
	\hspace{-1mm}
	\subfigure{\label{fault3-24}}\addtocounter{subfigure}{-2}
	\subfigure
	{\subfigure[Situation 24]{\includegraphics[width=0.236\textwidth]{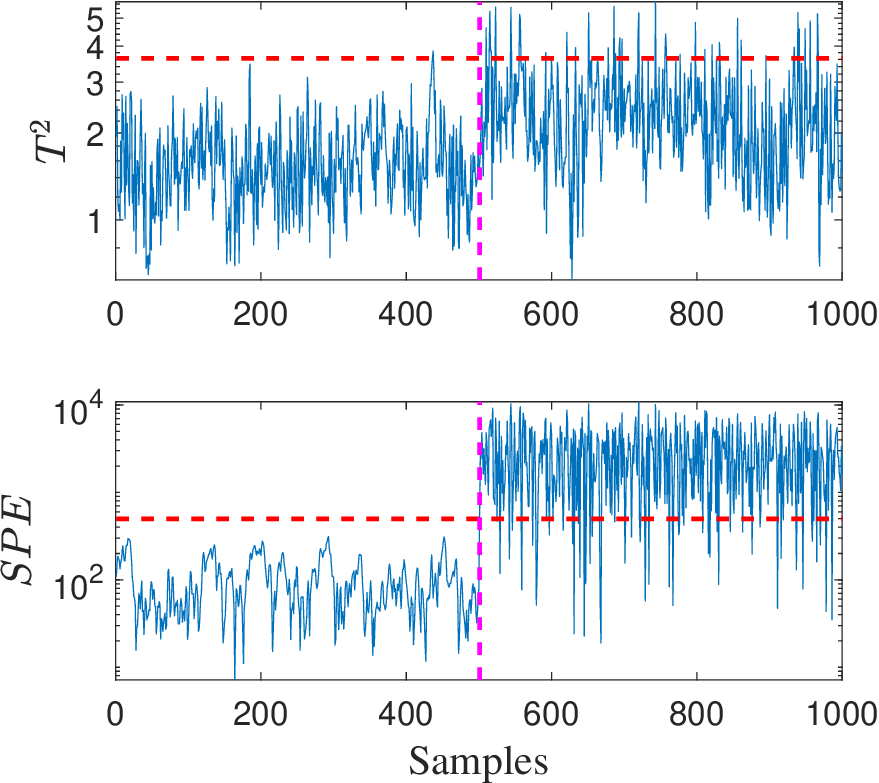}}}
	\vspace{-0.5mm}
	\hspace{-1mm}
	\subfigure{\label{fault3-25}}\addtocounter{subfigure}{-2}
	\subfigure
	{\subfigure[Situation 25]{\includegraphics[width=0.236\textwidth]{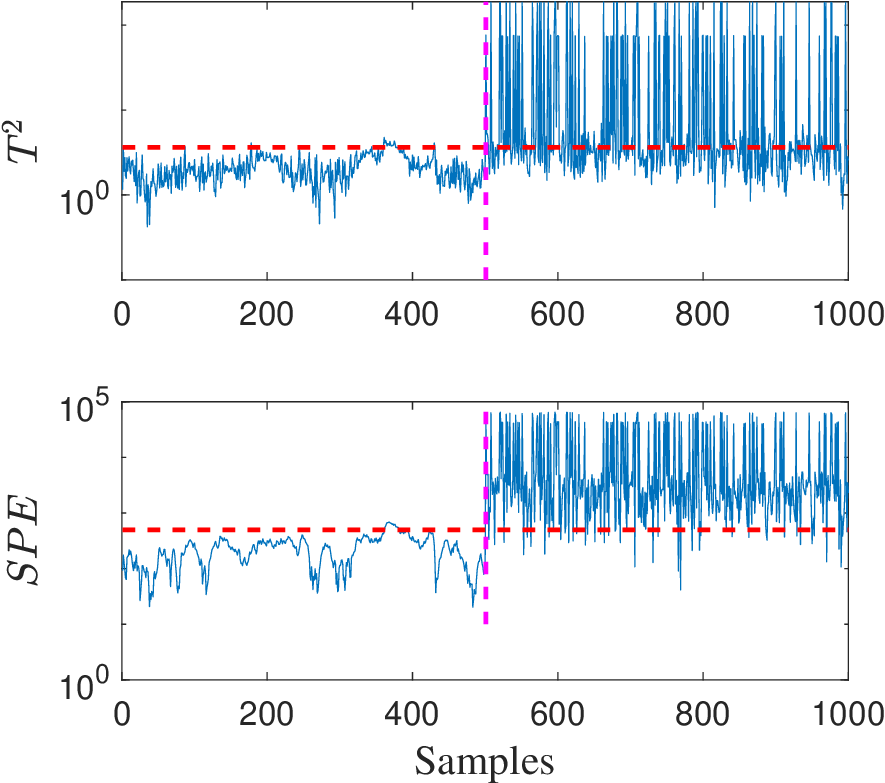}}}
	%
	%	\vspace{-0.5mm}
	%	\hspace{-1mm}
	%	\subfigure{\label{fault3-27}}\addtocounter{subfigure}{-2}
	%	\subfigure
	%	{\subfigure[Situation 27]{\includegraphics[width=0.236\textwidth]{GMMCVA_mode12_mode2_Fault1.eps}}}
	%
	\vspace{-0.5mm}
	\hspace{-1mm}
	\subfigure{\label{fault3-28}}\addtocounter{subfigure}{-2}
	\subfigure
	{\subfigure[Situation 28]{\includegraphics[width=0.236\textwidth]{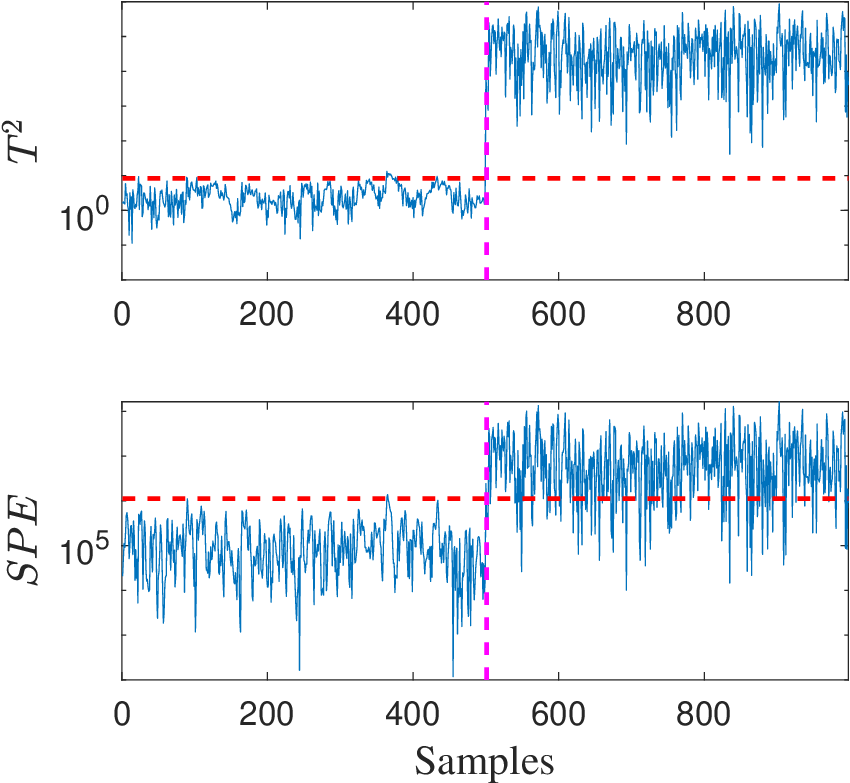}}}
	\vspace{-0.5mm}
	\hspace{-1mm}
	\subfigure{\label{fault3-29}}\addtocounter{subfigure}{-2}
	\subfigure
	{\subfigure[Situation 29]{\includegraphics[width=0.236\textwidth]{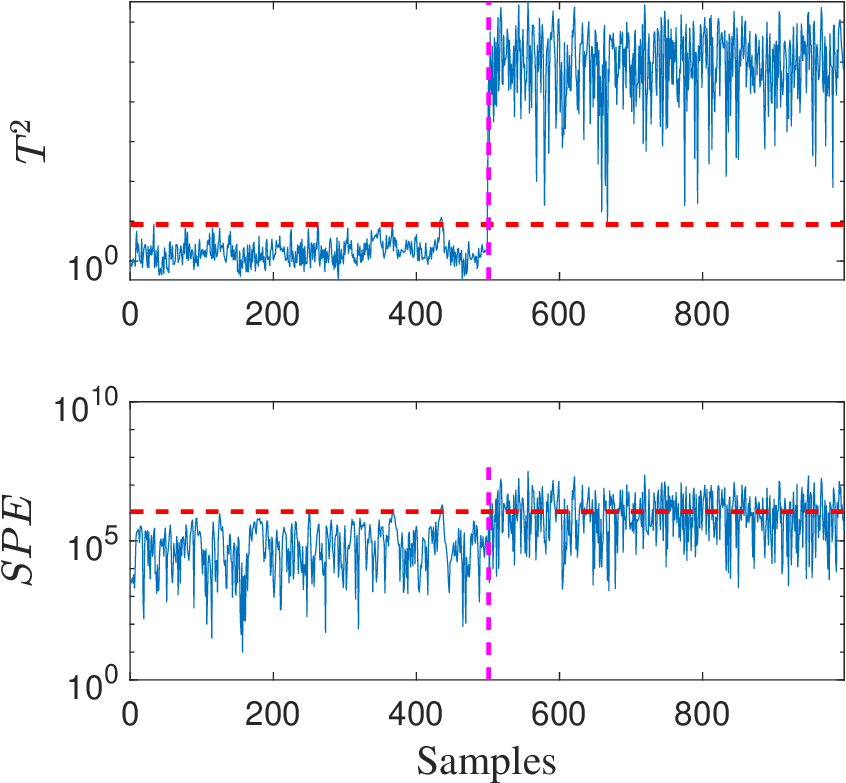}}}
	\vspace{-0.5mm}
	\hspace{-1mm}
	\subfigure{\label{fault3-30}}\addtocounter{subfigure}{-2}
	\subfigure
	{\subfigure[Situation 30]{\includegraphics[width=0.236\textwidth]{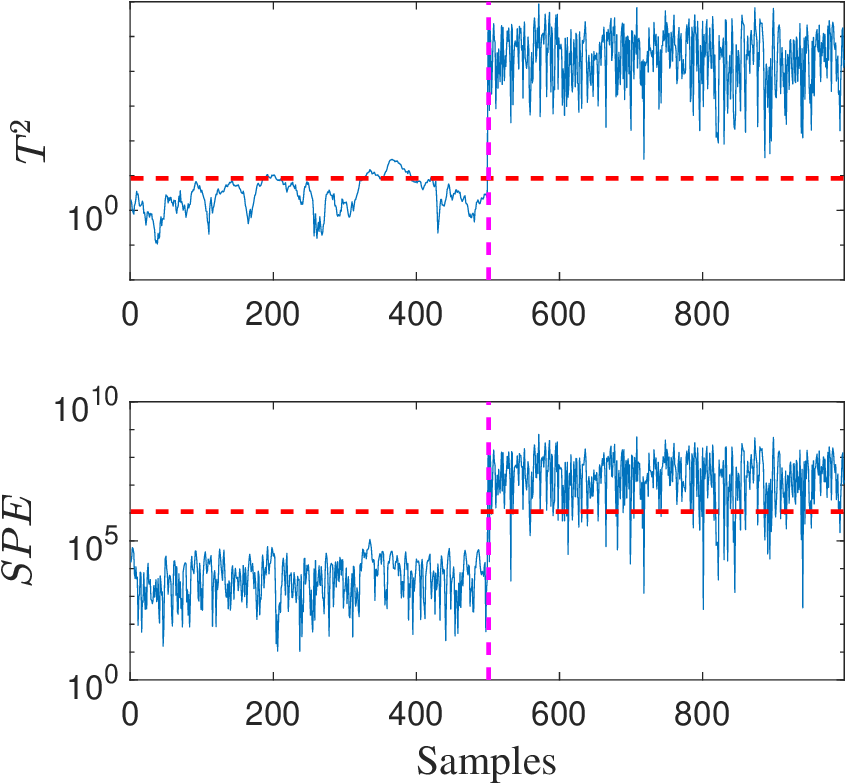}}}
	\centering
	\caption{Monitoring charts of the CSTH case} \label{case1}
	%\vspace{-1em}
\end{figure*}

	The monitoring results are summarized in Tables \ref{Table-PSFAEWC} and \ref{Table2-comparative}.  Partial monitoring charts  are depicted in Fig. \ref{case1}
	 owing to paper length. Generally, PSFA--EWC provides excellent performance for sequential modes, where the real fault and normal variations can be distinguished by $S^2$ statistic.  The FDRs of  $S^2$ statistic are higher than $98\%$, which indicate that the fault is detected accurately by PSFA--EWC. For PSFA, the FDRs of $S^2$ statistic are lower than $84\%$, especially for Situations 4 and 5. The FDRs of $T^2$  or $ SPE $ are similar for Situations 1--11. However, the FARs of Situation 11 are higher than $70\%$.  When a new mode is identified, a small number of normal samples are collected and utilized to  update the monitoring model. New features are extracted while consolidating the previously learned knowledge, which are sufficient to deliver excellent performance for the existing modes. Succinctly, PSFA--EWC can transfer knowledge between modes,  while it is difficult to establish an accurate PSFA model based on limited data.

	For Situations 12--14, RSFA fails to monitor the successive modes based on an adaptive model, where the FDRs of $S^2$ are less than $60\%$. RSFA is difficult to track the dramatic variations on the entire dataset.
%But the FARs of RSFA are pretty low. 
Analogous to PSFA--EWC, PCA--EWC is expected to offer prominent performance for sequential modes. However, the FDRs of Situations 16--20 cannot compare with the corresponding situations of PSFA--EWC. Although both methods utilize EWC to preserve the previously  learned knowledge, PSFA can deal with dynamic slow features and  $S^2$ is designed to reflect the unusual dynamic behaviors, while PCA is suitable to stationary data in each mode.
IMPPCA and MCVA build the local  models in each mode and the model needs to be retrained on the entire dataset when a new mode arrives. They deliver outstanding monitoring consequences  for the learned modes, expect for Situation 30. 
	
	Generally, PSFA--EWC outperforms others for sequential modes, where the number of modes and samples per mode are not required in advance.   When a new mode is identified, a few data are collected and the model is rapidly updated by assimilating new information while consolidating the learned knowledge. 
	 The RSFA model is updated when a new normal sample arrives, but fails to distinguish the normal changes and real faults in multimode nonstationary processes. Compared with PSFA--EWC, PCA--EWC is effective to detect the abnormality from static features but  difficult to identify a new mode. PSFA--EWC, RSFA and PCA--EWC have the basically fixed model capacity, where a single model is updated continually. For IMPPCA and MCVA, % identify the modes and build the local models within each mode. 
	 the model is rebuilt based on the entire dataset when a new mode arrives and the model complexity would increase with the emergence of novel  modes.

\begin{table*}[!tbp]
	\begin{center}
		\caption{FDRs ($\%$) and FARs ($\%$) for PCA--EWC, IMPPCA and MCVA}\label{Table2-comparative}
		\footnotesize
		%\small
		\renewcommand\arraystretch{1}
		\begin{tabular}{l  c c c c c     c c c c  }
			\hline
			%	\multirow{3}*{\makecell{Fault\\ type}} 
			&\multirow{3}*{Methods}  
			&\multicolumn{4}{c}{CSTH}     &\multicolumn{4}{c}{Pulverizing system}\\
			\cmidrule(r){3-6} \cmidrule(r){7-10} %\cmidrule(r){11-14}
			&	& \multicolumn{2}{c}{$T^2$}  & \multicolumn{2}{c}{$ SPE $}   & \multicolumn{2}{c}{$T^2$}  & \multicolumn{2}{c}{$ SPE $}      \\	
			\cmidrule(r){3-4}  \cmidrule(r){5-6}  \cmidrule(r){7-8}  \cmidrule(r){9-10}  
			&	& FDR & FAR    &FDR  & FAR          &FDR & FAR       & FDR &FAR   \\		    
			\hline			  
			Situation 15  &PCA       &83.0 &4.4     &94.2  &0.8       &99.92  &1.00   &99.92 &1.00  \\
			Situation 16  &PCA--EWC  &9.2  &0.6     &90.8  &0         &100  &0        &100 &0.16 \\
			Situation 17  &PCA--EWC  &8.2  &0.8     &91.6  &0         &99.92  &0.14   &99.92 &0\\
			Situation 18  &PCA--EWC  &13.0  &0.8    &90.6  &0         &100   &2.44    &100 &0.20\\
			Situation 19  &PCA--EWC  &12.2  &1.4    &92.4  & 0        &99.92  &0.14   &99.92 &0 \\
			Situation 20  &PCA--EWC  &12.6   &1.2   &91.8  &0         &99.45  &0      &99.45 &0 \\
			\hline
			Situation 21 &IMPPCA   &24.4  &0.8    &94.8   &0.4       &99.92  &0       &99.84 &0.29\\
			Situation 22 &IMPPCA   &0.8  &0.2     &91.0  &0          &99.45  &0.48    &99.45 &4.23\\
			Situation 23 &IMPPCA   &20.8  &0.8    &94.8  &0.4        &99.92  &0.29    &99.92 &0 \\
			Situation 24 &IMPPCA   &8.4  &0.2     &89.0  &0          &99.45  &0       &99.45 &0 \\
			Situation 25 &IMPPCA   &43.0 &5.2     &91.0  &4.2        &100  &63.96     &95.16 &47.72 \\
			\hline
			Situation 26 &MCVA    &100 &3.0      &77.71 &0.2        &100  &0.14     &0 &0\\
			Situation 27 &MCVA    &99.8 &1.6     &39.56 &0.2        &49.27  &0     &97.50 &0 \\
			Situation 28 &MCVA    &100  &3.0     &80.92 &0.6        &100  &0.14     &96.07 &0\\
			Situation 29 &MCVA    &99.8 &1.2     &49.20 &1.0        &100  &6.07     &99.63 &0\\
			Situation 30 &MCVA    &100 & 16.8    &87.75 &0.2        &100  &1.62     &97.34 &0\\
			\hline
		\end{tabular}
	\end{center}
\end{table*}

\begin{table}[!tbp]	
	\begin{center}
		\caption{Experimental data of the practical coal pulverizing system}\label{Table_informationZS}
		\footnotesize
		\renewcommand\arraystretch{1}
		\setlength\tabcolsep{3.5pt}
		\begin{tabular}{c c c  c l} %{c c c c}
			\hline
			\makecell{Mode\\ number} & \makecell{NoTrS} & \makecell{NoTeS} &\makecell{Fault\\ location}  &Fault cause  \\
			\hline  % & Yinni   Yinni   Aomeng 
			%\multirow{3}{*} {\makecell{Case \\3}} &\multirow{3}{4.2cm}{9 variables:   pressure of air powder mixture, outlet temperature,  primary air pressure and  temperature,    etc.}
			\makecell{ ${\mathcal M}_1$}  & 2520 &1440 & 699     & Pulverizer deflagration\\
			\makecell{ ${\mathcal M}_2$}  & 540   &1800 & 1253  &Hot primary air electric damper failure \\
			\makecell{${\mathcal M}_3$}   &540 &1440 & 986       &Air leakage at primary air interface\\
			\hline
		\end{tabular}
	\end{center}
\end{table}
	
	\begin{figure}[!bp]
		\centering
		\includegraphics[width=0.27\textwidth]{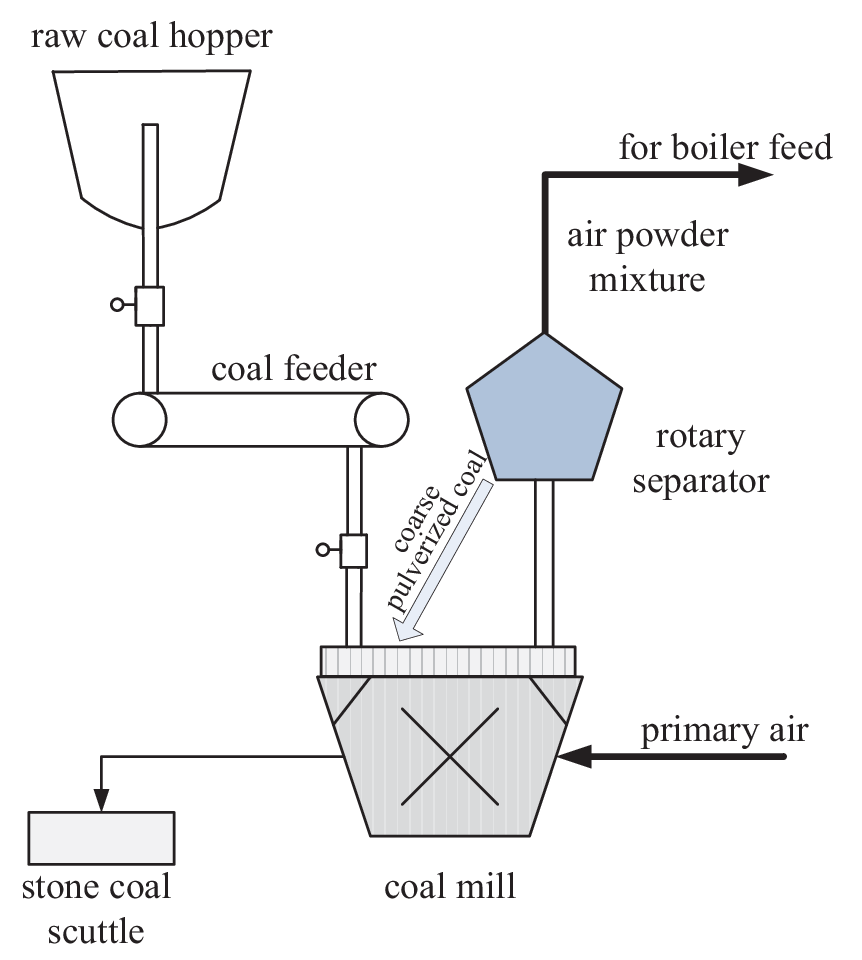}
		\caption{Schematic diagram of coal pulverizing system}
		\label{fig_benchmark}
	\end{figure}

	\subsection{The pulverizing system}
	
	We focus on  the coal pulverizing system of the 1030--MW ultra-supercritical thermal power plant in China \cite{zhang2021multimode}. 
	%Relevant introduction can refer to \cite{zhang2021multimode}. 
	The structure is depicted  in Fig. \ref{fig_benchmark}, which is composed of  coal feeder, coal mill, rotary separator, raw coal hopper and stone coal scuttle. 
	The coal pulverizing system grinds the raw coal into pulverized coal with desired  fineness and optimal temperature. 
	%High temperature may lead to deflagration and low temperature would reduce the combustion efficiency.  
	According to the historical recording, the fault in outlet temperature occurs frequently and it is essential to investigate this sort of fault. Data from three successive modes are selected to illustrate the effectiveness, as listed in Table \ref{Table_informationZS}. When a new mode arrives, only 540 normal samples are collected to update the model. The variables are selected by expert experience and prior knowledge.

\begin{figure*}[!tbp]
		\centering
		\subfigure{\label{fault2-2}}\addtocounter{subfigure}{-2}
		\subfigure
		{\subfigure[Situation 2]{\includegraphics[width=0.236\textwidth]{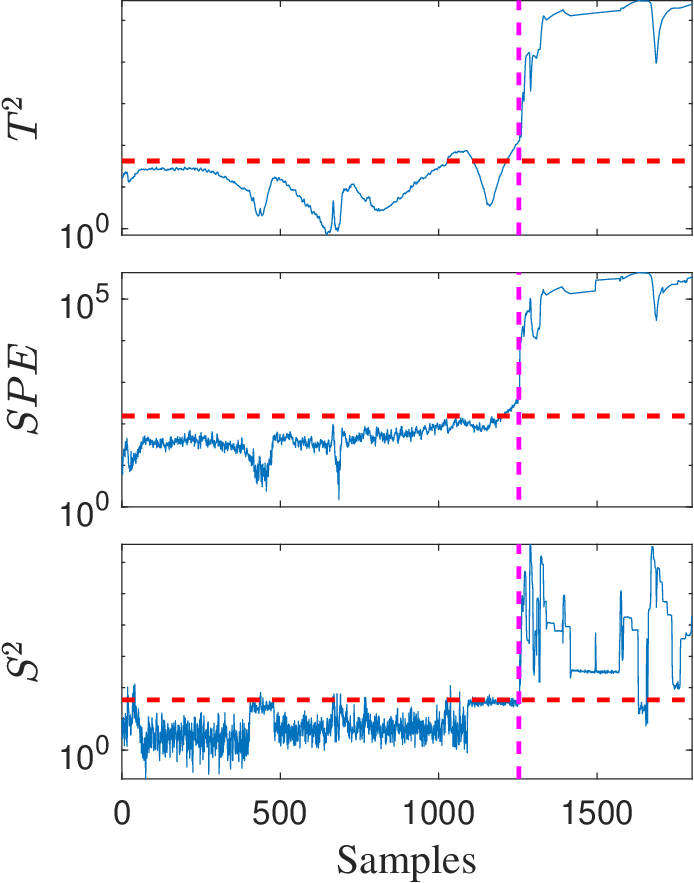}}}
		%	\vspace{-0.5mm}
		%	\hspace{-1mm}
		%	\subfigure{\label{fault2-3}}\addtocounter{subfigure}{-2}
		%	\subfigure
		%	{\subfigure[Situation 3]{\includegraphics[width=0.236\textwidth]{PSFAEWC_mode12_mode1_Fault3.eps}}}
		\vspace{-0.5mm}
		\hspace{-1mm}
		\subfigure{\label{fault2-4}}\addtocounter{subfigure}{-2}
		\subfigure
		{\subfigure[Situation 4]{\includegraphics[width=0.236\textwidth]{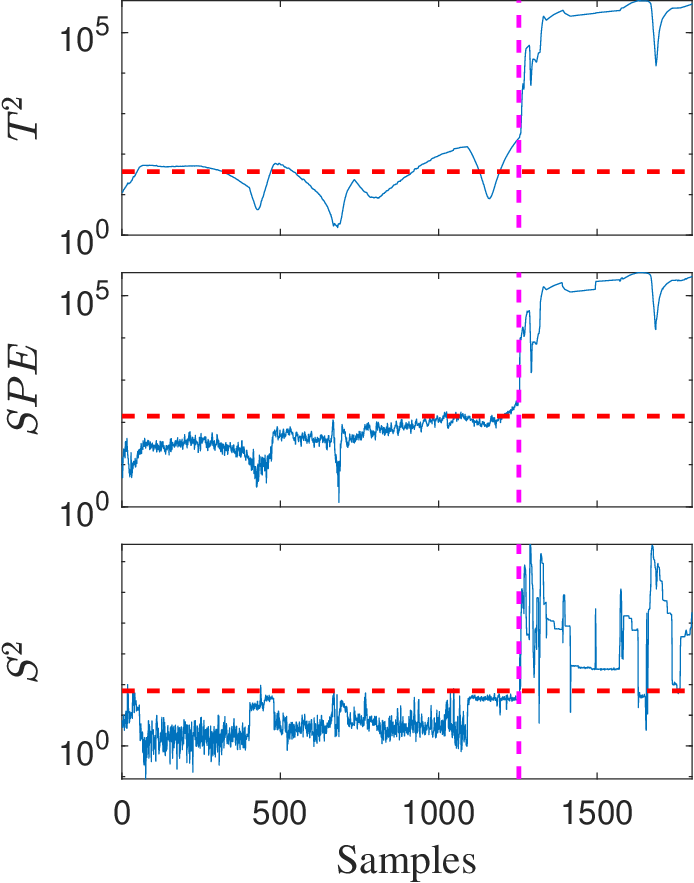}}}
		%	
		%	\vspace{-0.5mm}
		%	\hspace{-1mm}
		%	\subfigure{\label{fault2-5}}\addtocounter{subfigure}{-2}
		%	\subfigure
		%	{\subfigure[Situation 5]{\includegraphics[width=0.236\textwidth]{PSFAEWC_mode2_mode1_Fault3.eps}}}
		%
		\vspace{-0.5mm}
		\hspace{-1mm}
		\subfigure{\label{fault2-6}}\addtocounter{subfigure}{-2}
		\subfigure
		{\subfigure[Situation 6]{\includegraphics[width=0.236\textwidth]{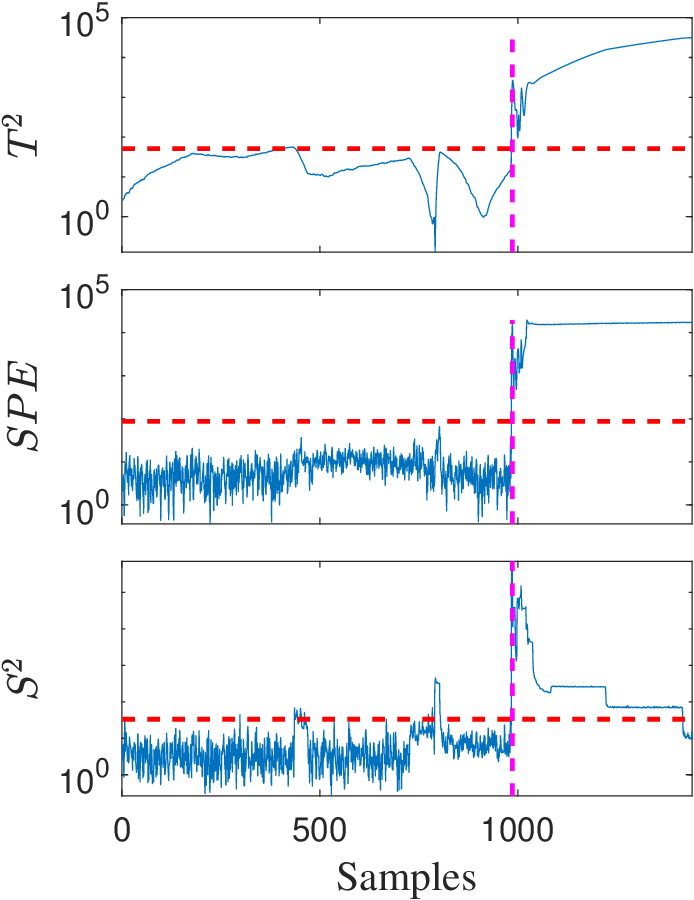}}}
		\vspace{-0.5mm}
		\hspace{-1mm}
		\subfigure{\label{fault2-7}}\addtocounter{subfigure}{-2}
		\subfigure
		{\subfigure[Situation 7]{\includegraphics[width=0.236\textwidth]{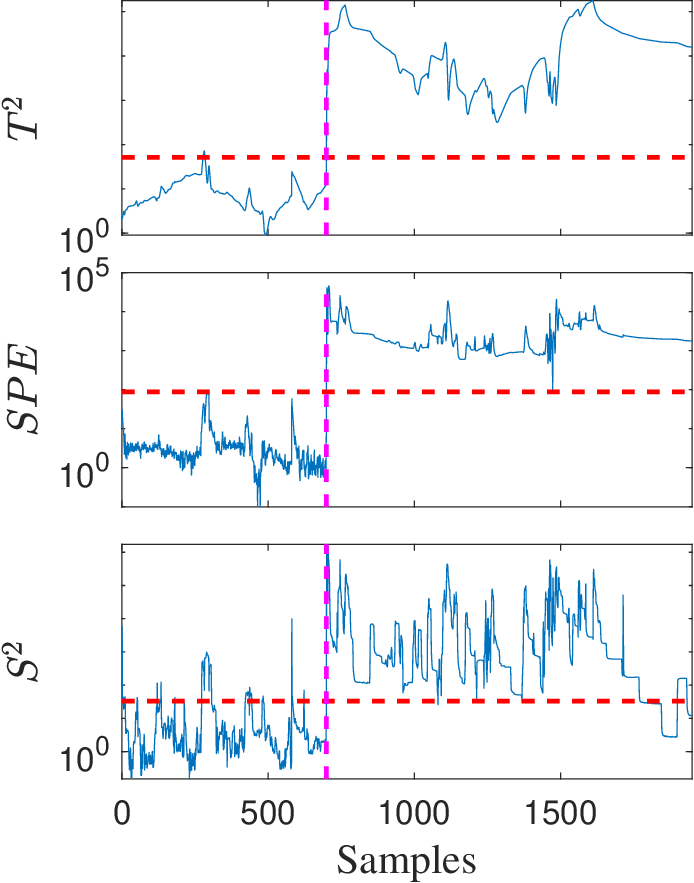}}}
		\vspace{-0.5mm}
		\hspace{-1mm}   	
		\subfigure{\label{fault2-8}}\addtocounter{subfigure}{-2}
		\subfigure
		{\subfigure[Situation 8]{\includegraphics[width=0.236\textwidth]{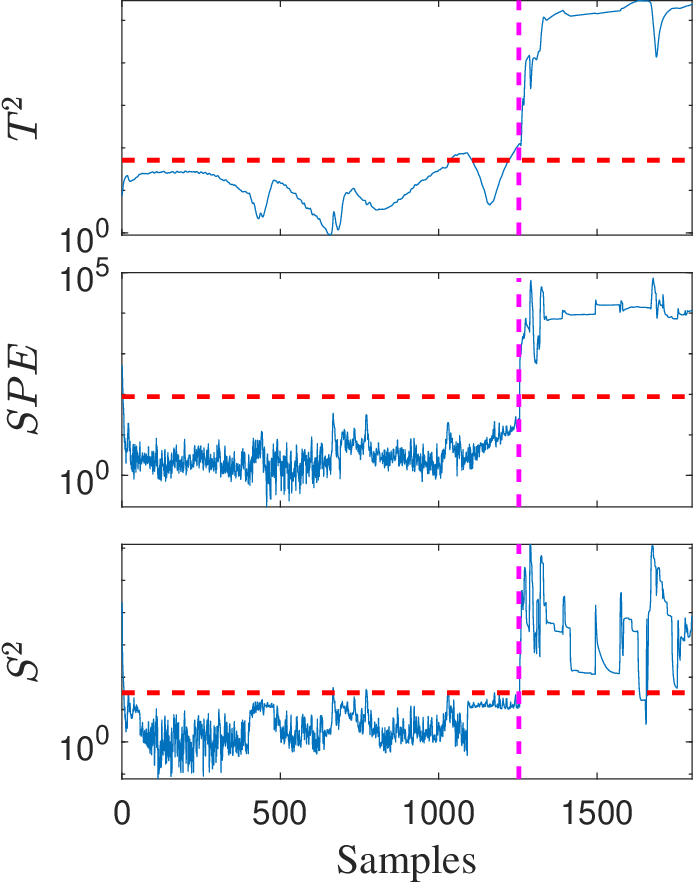}}}
		\vspace{-0.5mm}
		\hspace{-1mm}   	
		\subfigure{\label{fault2-9}}\addtocounter{subfigure}{-2}
		\subfigure
		{\subfigure[Situation 9]{\includegraphics[width=0.236\textwidth]{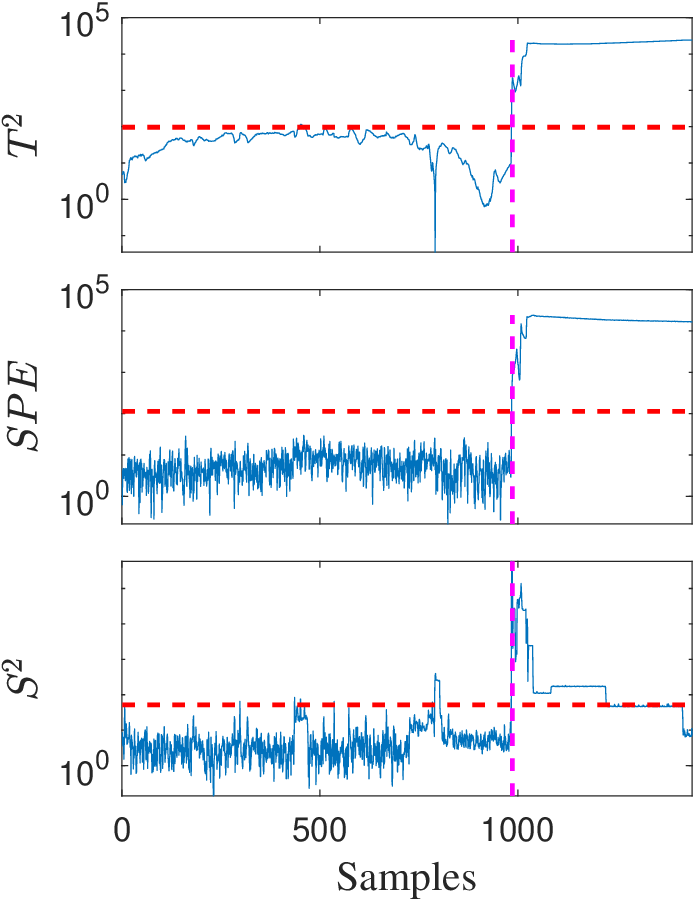}}}
		%
		%	\vspace{-0.5mm}
		%	\hspace{-1mm}
		%	\subfigure{\label{fault2-10}}\addtocounter{subfigure}{-2}
		%	\subfigure
		%	{\subfigure[Situation 10]{\includegraphics[width=0.236\textwidth]{PSFAEWC_mode3_mode1_Fault3.eps}}}
		\vspace{-0.5mm}
		\hspace{-1mm}
		\subfigure{\label{fault2-11}}\addtocounter{subfigure}{-2}
		\subfigure
		{\subfigure[Situation 11]{\includegraphics[width=0.236\textwidth]{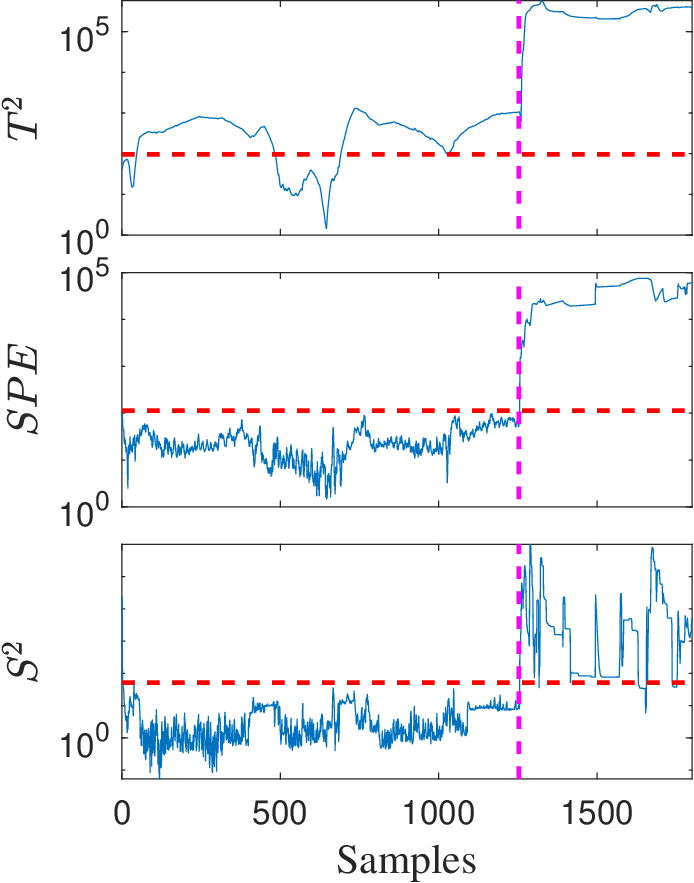}}}
		%	\vspace{-0.5mm}
		%	\hspace{-1mm}
		%	\subfigure{\label{fault2-13}}\addtocounter{subfigure}{-2}
		%	\subfigure
		%	{\subfigure[Situation 13]{\includegraphics[width=0.236\textwidth]{RSFA_mode12_mode2_Fault3.eps}}}
		\vspace{-0.5mm}
		\hspace{-1mm}
		\subfigure{\label{fault3-14}}\addtocounter{subfigure}{-2}
		\subfigure
		{\subfigure[Situation 14]{\includegraphics[width=0.236\textwidth]{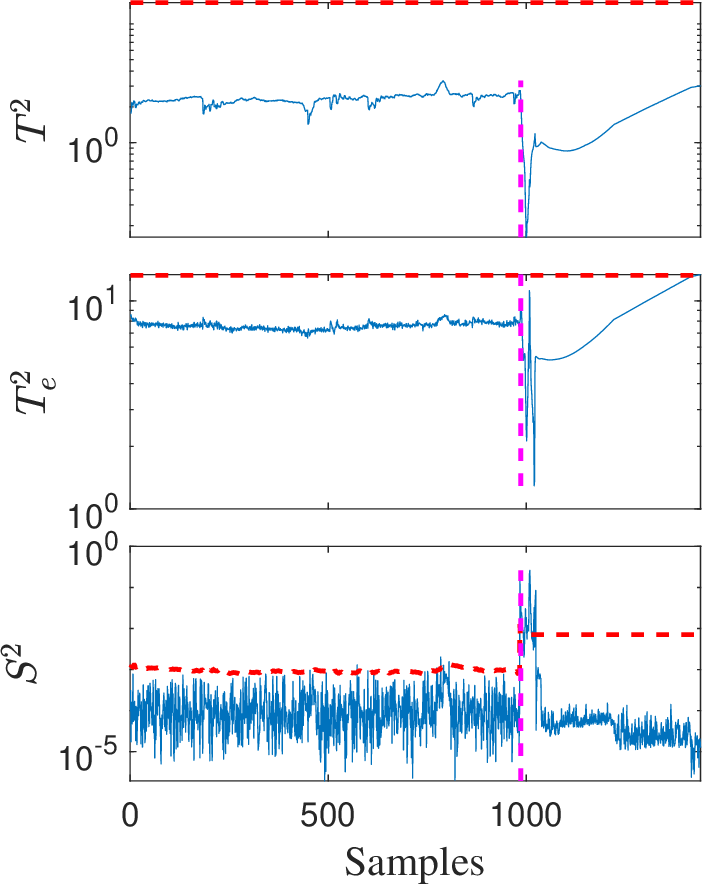}}}
		%
		%	\vspace{-0.5mm}
		%	\hspace{-1mm}
		%	\subfigure{\label{fault2-16}}\addtocounter{subfigure}{-2}
		%	\subfigure
		%	{\subfigure[Situation 16]{\includegraphics[width=0.236\textwidth]{PCAEWC_mode12_mode2_Fault3.eps}}}
		%
		%	\vspace{-0.5mm}
		%	\hspace{-1mm}
		%	\subfigure{\label{fault2-17}}\addtocounter{subfigure}{-2}
		%	\subfigure
		%	{\subfigure[Situation 17]{\includegraphics[width=0.236\textwidth]{PCAEWC_mode12_mode1_Fault3.eps}}}
		%
		\vspace{-0.5mm}
		\hspace{-1mm}
		\subfigure{\label{fault2-18}}\addtocounter{subfigure}{-2}
		\subfigure
		{\subfigure[Situation 18]{\includegraphics[width=0.236\textwidth]{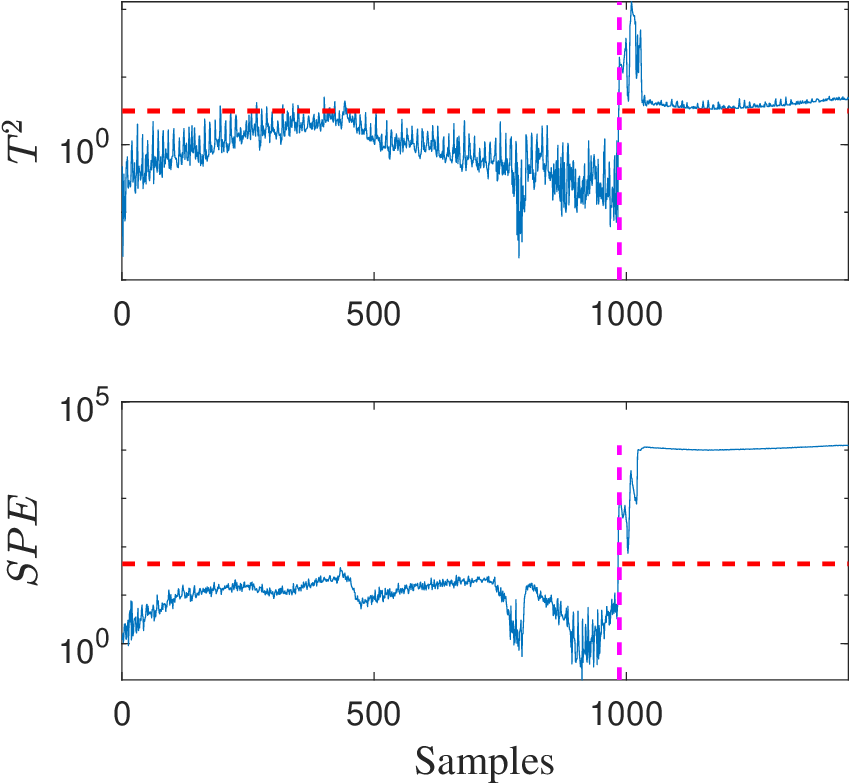}}}
		%
		%	\vspace{-0.5mm}
		%	\hspace{-1mm}
		%	\subfigure{\label{fault2-19}}\addtocounter{subfigure}{-2}
		%	\subfigure
		%	{\subfigure[Situation 19]{\includegraphics[width=0.236\textwidth]{PCAEWC_mode123_mode1_Fault3.eps}}}
		%
		\vspace{-0.5mm}
		\hspace{-1mm}
		\subfigure{\label{fault2-20}}\addtocounter{subfigure}{-2}
		\subfigure
		{\subfigure[Situation 20]{\includegraphics[width=0.236\textwidth]{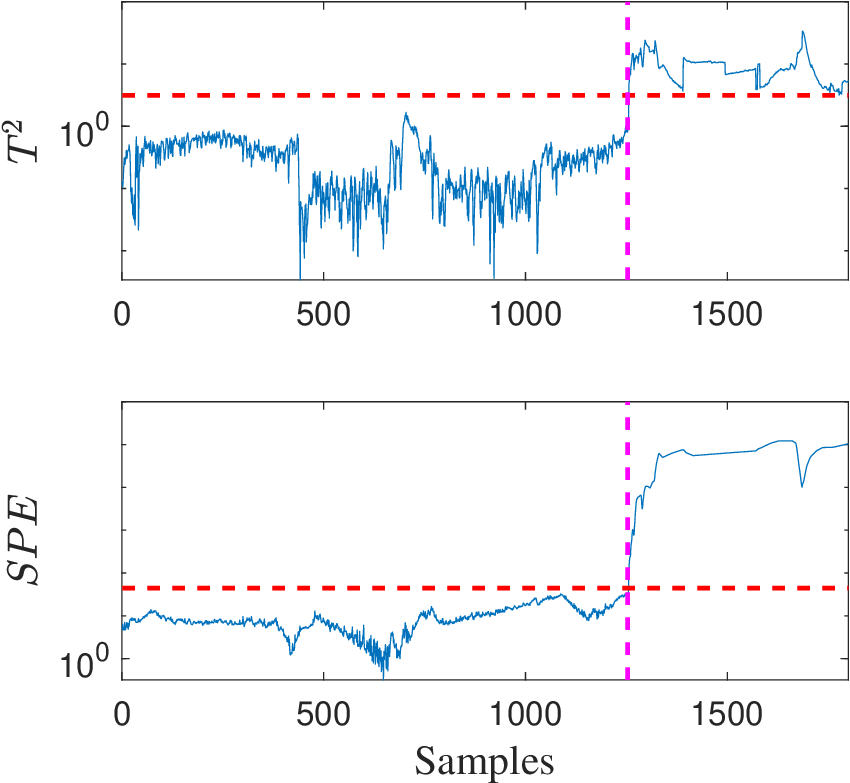}}}
		%
		%	\vspace{-0.5mm}
		%	\hspace{-1mm}
		%	\subfigure{\label{fault2-22}}\addtocounter{subfigure}{-2}
		%	\subfigure
		%	{\subfigure[Situation 22]{\includegraphics[width=0.236\textwidth]{IMPPCA_mode12_mode2_Fault3fig2.eps}}}
		%
		\vspace{-0.5mm}
		\hspace{-1mm}
		\subfigure{\label{fault2-23}}\addtocounter{subfigure}{-2}
		\subfigure
		{\subfigure[Situation 23]{\includegraphics[width=0.236\textwidth]{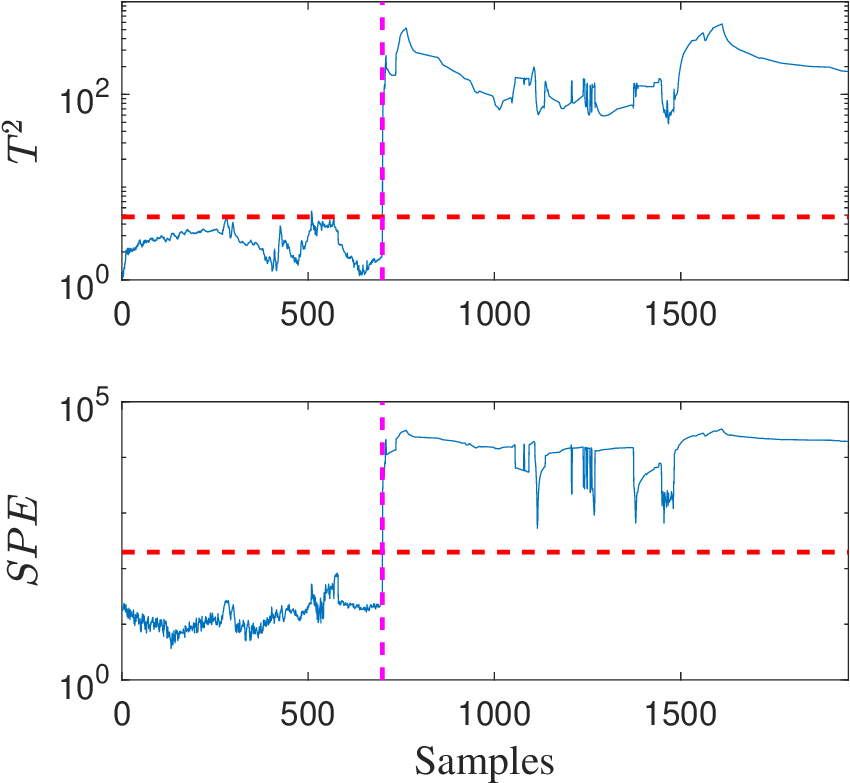}}}
		\vspace{-0.5mm}
		\hspace{-1mm}
		\subfigure{\label{fault2-24}}\addtocounter{subfigure}{-2}
		\subfigure
		{\subfigure[Situation 24]{\includegraphics[width=0.236\textwidth]{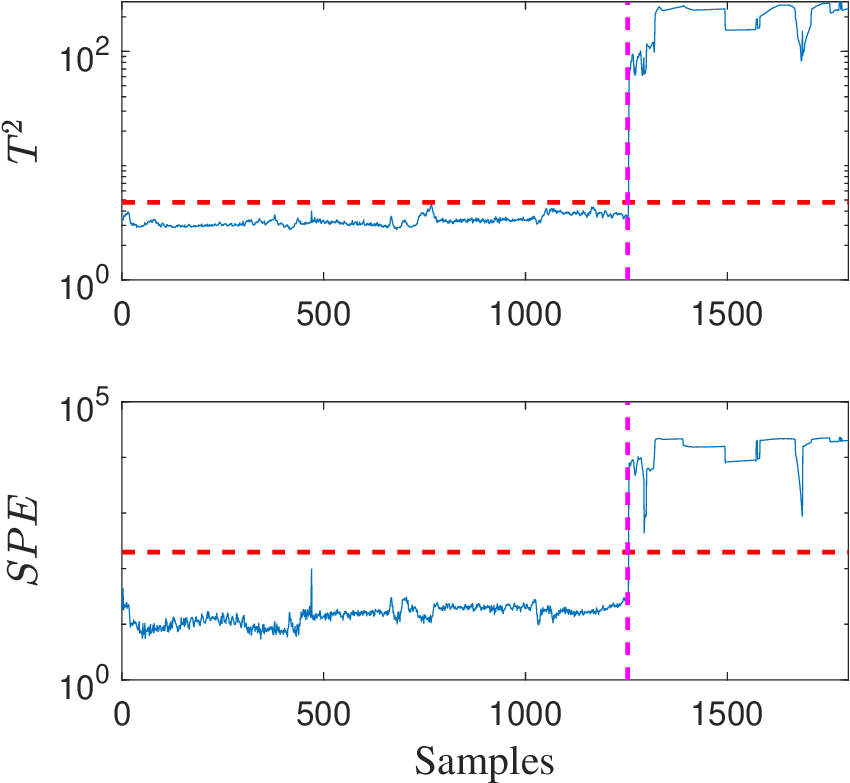}}}
		\vspace{-0.5mm}
		\hspace{-1mm}
		\subfigure{\label{fault2-25}}\addtocounter{subfigure}{-2}
		\subfigure
		{\subfigure[Situation 25]{\includegraphics[width=0.24\textwidth]{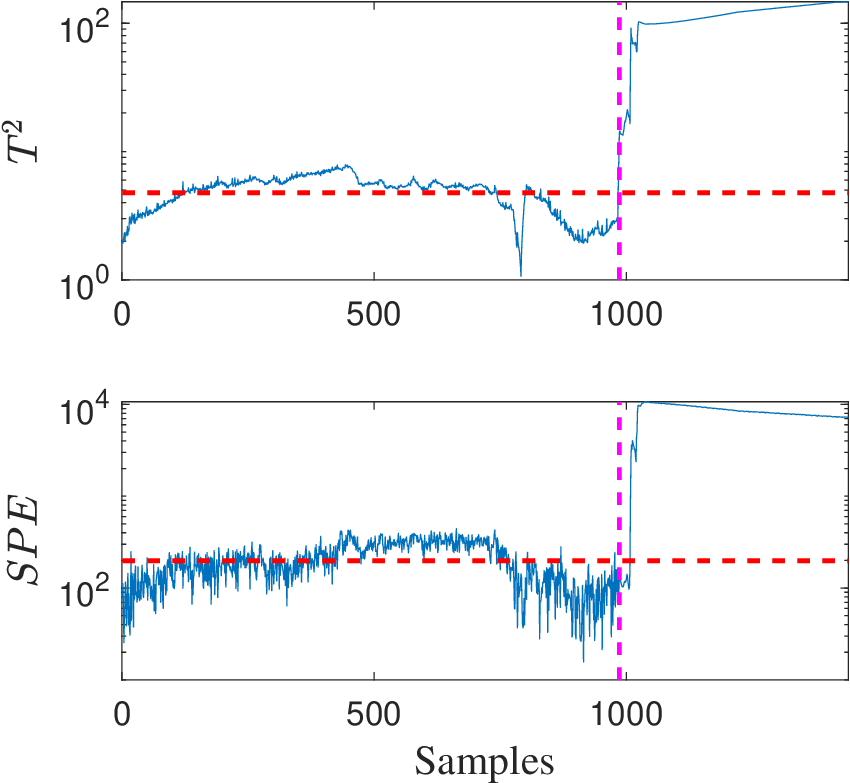}}}
		%
		%	\vspace{-0.5mm}
		%	\hspace{-1mm}
		%	\subfigure{\label{fault2-27}}\addtocounter{subfigure}{-2}
		%	\subfigure
		%	{\subfigure[Situation 27]{\includegraphics[width=0.236\textwidth]{GMMCVA_mode12_mode2_Fault3.eps}}}
		%
		\vspace{-0.5mm}
		\hspace{-1mm}
		\subfigure{\label{fault2-28}}\addtocounter{subfigure}{-2}
		\subfigure
		{\subfigure[Situation 28]{\includegraphics[width=0.236\textwidth]{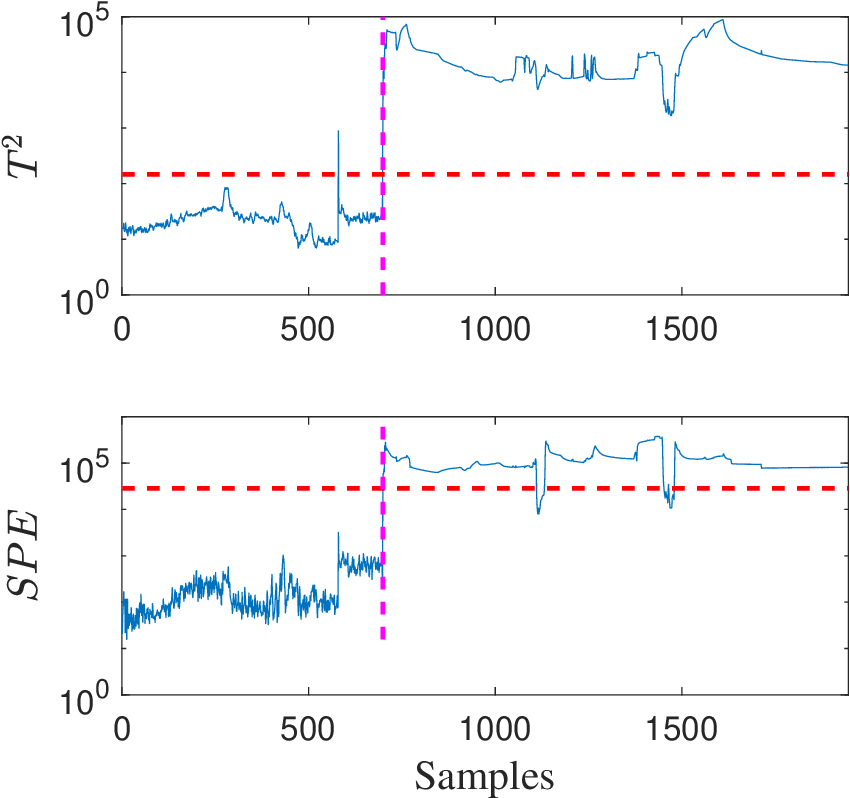}}}
		\vspace{-0.5mm}
		\hspace{-1mm}
		\subfigure{\label{fault2-29}}\addtocounter{subfigure}{-2}
		\subfigure
		{\subfigure[Situation 29]{\includegraphics[width=0.236\textwidth]{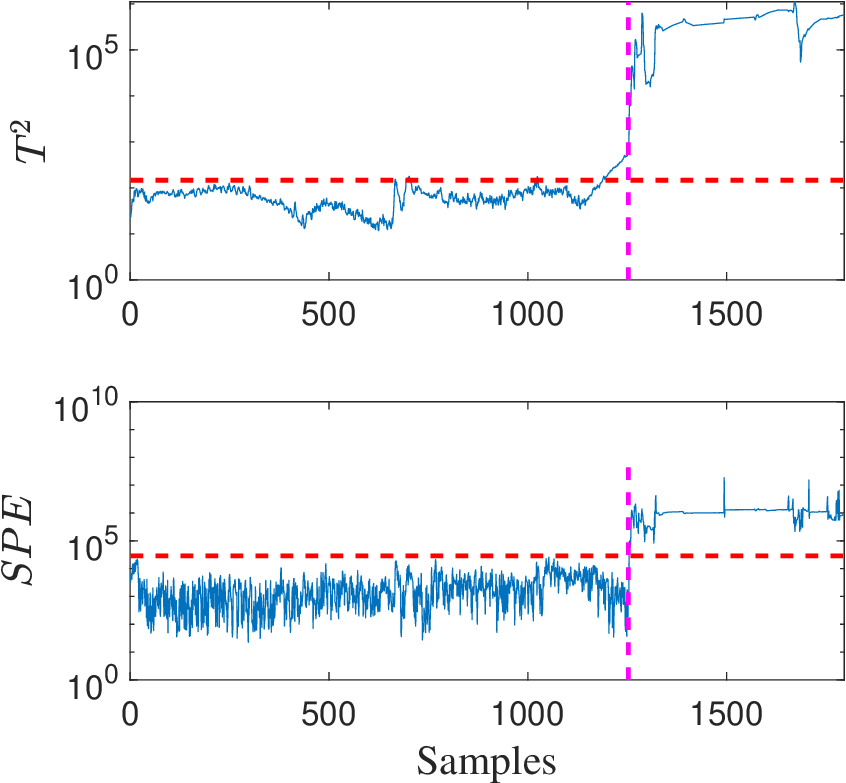}}}
		\vspace{-0.5mm}
		\hspace{-1mm}
		\subfigure{\label{fault2-30}}\addtocounter{subfigure}{-2}
		\subfigure
		{\subfigure[Situation 30]{\includegraphics[width=0.234\textwidth]{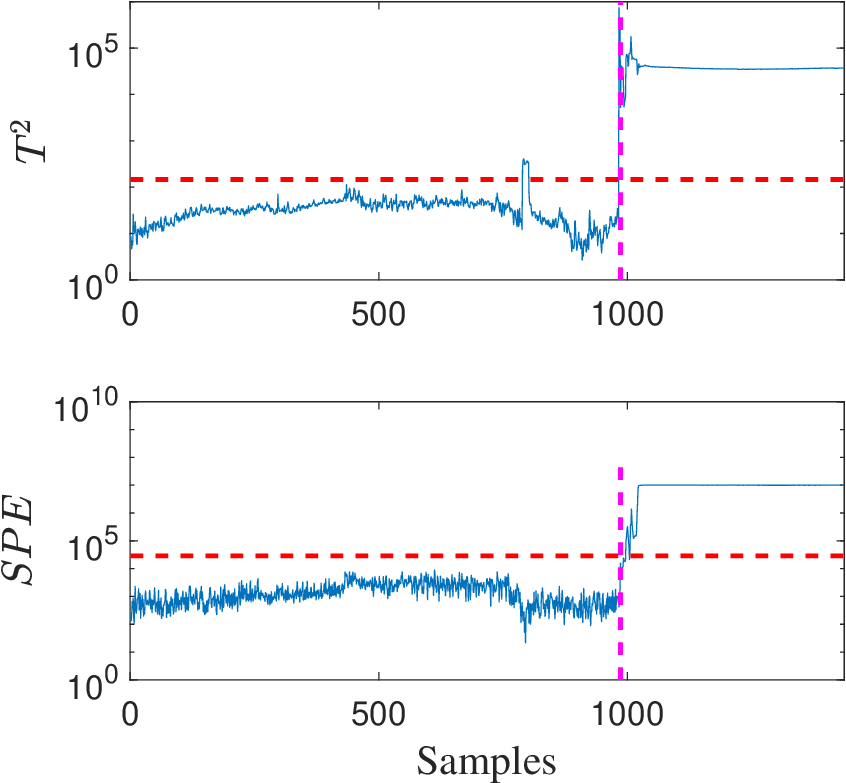}}}
		\centering
		\caption{Monitoring charts of the pulverizing system} \label{case2}
		%\vspace{-1em}
	\end{figure*}

	The simulation results of 30 situations are summarized in Table \ref{Table-PSFAEWC} and Table \ref{Table2-comparative}. Partial monitoring charts are described in Fig. \ref{case2}. 
	%PSFA--EWC performs better than PSFA and the significant information preserved in the model is valuable to enhance detection accuracy. 
	With regard to $S^2$ statistic, the FDRs of Situations 6 and 9 are $94.73\%$ and $53.63\%$. which indicates that the perviously learned knowledge from modes $\mathcal{M}_1$ and $\mathcal{M}_2$  is conducive to monitor mode $\mathcal{M}_3$. For $T^2$ statistic, the FARs of Situations 4, 10 and 11 are higher than $48\%$, while the FARs of PSFA--EWC are lower than $10\%$.  PSFA suffers from catastrophic forgetting issue, where the model for one mode may not provide excellent performance for  another mode.
	%  PSFA--EWC provides excellent performance than PSFA.
	%The FDRs of Situation 3 are better than those of Situation 1, which indicates that the learned knowledge of mode  $\mathcal{M}_2$  is conducive to monitor mode $\mathcal{M}_1$. 	%The FARs of Situations 5, 10 and 11 are higher than $20\%$, which are not acceptable. 
	%PSFA suffers from catastrophic forgetting issue, where the model for one mode may not provide excellent performance for  another mode.  
	RSFA cannot monitor the multiple modes accurately and the FDRs of $S^2$ are lower than $44\%$. Only the FDR of $T^2$ is $84.49\%$ for Situation 13. PCA--EWC can detect the faults in successive modes timely and the FDRs approach $100\%$. IMPPCA and MCVA  offer favorable monitoring performance and the FDRs are convincing, except for Situation 25.

	In conclusion, PSFA--EWC is capable of monitoring sequential modes accurately and the fault is confirmed by  $S^2$ statistic.  The model is updated continually by extracting new information while preserving the learned knowledge, thus avoiding performance degradation  for similar modes as before. RSFA is effective to deal with slowly time-varying data and thus fails to track the dramatic changes on the entire dataset. Similar to PSFA--EWC, PCA--EWC enables to monitor the multiple modes based on a updated model for this case. IMPPCA and MCVA are able to monitor the learned modes.
	%However, they require  the number of modes is a prior before learning and local models are retrained when new modes arrive. 
	In terms of detection accuracy, the model complexity and applications,  PSFA--EWC is the most desirable among five typical methods.
	%PCA-EWC 
	
	\section{Conclusion}\label{sec5}
	%This paper has introduced a novel extension of elastic weight consolidation to  multimode dynamic process monitoring, based on  multimode PSFA.   
	This paper has introduced a multimode PSFA algorithm with continual learning ability for multimode nonstationary process monitoring.
	The proposed PSFA--EWC method has powerful probabilistic interpretability and ability to deal with the measurement noise. When a new mode arrives, assume that a small set of data are collected, the single model is updated by consolidating new information while preserving the learned features. 
	The previously learned knowledge is retained and may be beneficial to establishing an accurate model for future relevant modes, thus delivering backward  and forward transfer learning ability. 
	%The learned knowledge can be transferred and the significant information of one mode may be  valuable to build an accurate model for another mode. 
	The PSFA features are extracted to form meaningful statistics for fault detection covering multimodes with only using recent mode data, with low storage and computational costs.  Compared with several state-of-the-art methods, the effectiveness of PSFA--EWC is illustrated by a CSTH case and a practical coal pulverizing system. 
	
%	In future work, we would investigate the multimode dynamic modes with applications to  chemical systems, industrial manufacturing systems, etc. Besides, the modes are diverse and the long-term continual learning ability is desired.
%	
%	 \cite{martens2020new}
	\appendix
\subsection{Estimation of Fisher information matrix with PSFA}\label{appendix:FIM}
In order to approximate the posterior probability  $P\left( \theta |{\cal M} _{i=1}^{K-1}  \right)$, sequentially with incoming modes $K=2,\dots $,  Laplacian approximation~\cite{rahaf2019continual,martens2020new} is employed, i.e., local Gaussian probability density is used for its approximation centered at maximum posterior probability
$\theta _{{\mathcal{M}_{K- 1}}}^*$, with covariance of the gradient of the model's log likelihood function with respect to  $\theta _{{\mathcal{M}_{K- 1}}}^*$.  
The Fisher information matrix  is 	the covariance of the gradient of the model's log likelihood function with respect to  the local optimum,  namely,
%which is approximated by
%. The empirical Fisher information is used as the approximation of true Fisher information matrix and can be calculated by
\begin{equation}\label{estimation_FIM}
	\begin{aligned}
		{\boldsymbol F}  
		& = {\mathbb E_{{P_{\boldsymbol x,\boldsymbol y}}}}\left[ {\nabla \log \;P\left( {\boldsymbol x,\boldsymbol y|\theta } \right)\nabla \log \;P{{\left( {\boldsymbol x,\boldsymbol y|\theta } \right)}^T}} \right]\\
		%	& = \sum\limits_t {p\left( {{\boldsymbol x_t},{\boldsymbol y_t}} \right)\nabla \log \;p\left( {{\boldsymbol x_t},{\boldsymbol y_t}|\theta } \right)\nabla \log \;p{{\left( {{\boldsymbol x_t},{\boldsymbol y_t}|\theta } \right)}^T}} 
		& = \frac{1}{T}\sum\limits_t \left[ {\nabla \log \;P\left( {\boldsymbol x_t,\boldsymbol y_t|\theta } \right)\nabla \log \;P{{\left( {\boldsymbol x_t,\boldsymbol y_t|\theta } \right)}^T}} \right]
	\end{aligned}
\end{equation}	
%	\end{equation}
where $\theta = \theta _{{\mathcal{M}_{K- 1}}}^*$ after the mode $\mathcal{M}_{K-1}$ has been learned. 	
The conditional probability is calculated by
%	\begin{equation}
\begin{center}
	$ P\left( {{\boldsymbol x_t},{\boldsymbol y_t}|\theta } \right) = P\left( {{\boldsymbol x_t}|{\boldsymbol y_t},{\theta _x}} \right)P\left( {{\boldsymbol y_t}|\theta_y} \right)$
\end{center}
%	\end{equation}

Within the context of our PSFA model (13), parameters $\boldsymbol V$ and $\boldsymbol \Lambda$ are considered  to calculate  the corresponding Fisher information matrices, since the Laplacian is based on well-behaved function approximation which may not be applicable to noise. Besides, it is reasonable to assume that variance of unknown noise is constant in our problem.    
The gradient with regard to $ \boldsymbol V $  is 
\begin{equation}\label{graf_V}
	\begin{aligned}
		%	\frac{{\;\partial \log \;p\left( {{x_t},{y_t}|\theta } \right)}}{{\partial V}} 
		{\nabla _{\boldsymbol V}}\log \;P\left( {\boldsymbol x_t,\boldsymbol y_t|\theta } \right)
		&= \frac{{\;\partial \log \;P\left( {{\boldsymbol x_t}|{\boldsymbol y_t},{\theta _x}} \right)}}{{\partial {\boldsymbol V}}}\\
		&	=\boldsymbol \Sigma _x^{ - 1}\left( {\boldsymbol V{\boldsymbol y_t} - {\boldsymbol x_t}} \right) \boldsymbol y_t^T
	\end{aligned}
\end{equation}
%	According to (\ref{joint_probab}) and (\ref{graf_V}),  
When the mode  $\mathcal{M}_K$ has been learned, the Fisher information matrix about $ \boldsymbol V $ is computed by
\begin{equation}\label{fisher_V}
	\begin{aligned}
		&{ {\boldsymbol F}}^V_{\mathcal{M}_K}  \\
		= &\frac{1}{T_K} \sum\limits_t { \boldsymbol \Sigma _x^{ - 1}\left( {\boldsymbol V_{\mathcal{M}_K}{\boldsymbol y_t} - {\boldsymbol x_t}} \right) \boldsymbol y_t^T \boldsymbol y_t \left( {\boldsymbol V_{\mathcal{M}_K}{\boldsymbol y_t} - {\boldsymbol x_t}} \right)^T \boldsymbol \Sigma _x^{ - 1}} 
	\end{aligned}
\end{equation}

Since $ \boldsymbol \Sigma = \boldsymbol I- {\boldsymbol  \Lambda}^2 $, $ {\boldsymbol y_t} \sim N\left( {\boldsymbol \Lambda {\boldsymbol y_{t - 1}}, \boldsymbol \Sigma } \right) $, 
and 
$\boldsymbol  \Lambda = diag\left( {{\lambda _1}, \ldots ,{\lambda _p}} \right)$,  
the gradient with respect to $\lambda_i$ is 
\begin{equation}\label{grad_lam}
	\begin{aligned}
	 &	{\nabla _{{\lambda _i}}}\log \;P\left( {\boldsymbol x_t, \boldsymbol y_t|\theta } \right) \\
		= &\frac{{\;\partial \log \;P\left( {{\boldsymbol y_t}|{\theta _y}} \right)}}{{\partial {\lambda _i}}}\\
		= &\frac{{ - \lambda _i^3 + {y_{t,i}}{y_{t - 1,i}}\lambda _i^2 + \left( {1 - y_{t,i}^2 - y_{t - 1,i}^2} \right){\lambda _i} + {y_{t,i}}{y_{t - 1,i}}}}{{{{\left( {1 - \lambda _i^2} \right)}^2}}}\\
		\triangleq & g\left(y_{t,i}, y_{t-1,i}, \lambda_i  \right)
	\end{aligned}
\end{equation}
For mode $\mathcal{M}_K$, the Fisher information matrix about $\lambda_i$ is 
\begin{equation}\label{fisher_lambda}
	F_{\lambda_i}=\frac{1}{T_K}\sum\limits_t {g\left(y_{t,i}, y_{t-1,i}, \lambda_{\mathcal{M}_K,i}  \right)^2}, i=1,\cdots,p 
\end{equation}
where $ \lambda_{\mathcal{M}_K,i} $ is the $i$th element of diagonal matrix ${\boldsymbol \Lambda}_{\mathcal{M}_K}$,  
${\mathcal {\boldsymbol F}}^{\Lambda}_{\mathcal{M}_K}  = diag \left({ F}_{\lambda_1},\cdots,{ F}_{\lambda_p} \right)$. 

\subsection{Estimation of sufficient statistics}\label{sufficient}

Similar to \cite{zafeiriou2016probabilistic}, Kalman filter  and Tanch-Tung-Striebel (RTS) smoother \cite{sarkka2008unscented} are adopted,  which contains the forward and backward recursion steps. 

First,  the forward recursions are adopted to  estimate the posterior distribution
$P\left( {{\boldsymbol  y_t}|{\boldsymbol  x_1},{\boldsymbol  x_2}, \cdots ,{\boldsymbol  x_t},{\theta ^{old}}} \right) \sim N\left( {{\boldsymbol  \mu _t},{\boldsymbol  U_t}} \right)$ sequentially. The posterior marginal is calculated by
%\begin{equation}
\begin{center}
	$		\begin{aligned}
		&\int {N\left( {{\boldsymbol y_{t - 1}}|{\boldsymbol \mu _{t - 1}},{\boldsymbol U_{t - 1}}} \right)} N\left( {{\boldsymbol y_t}|\boldsymbol \Lambda {\boldsymbol y_{t - 1}},\boldsymbol \Sigma } \right)d{\boldsymbol y_{t - 1}} \\
		&= N\left( {{\boldsymbol y_t}|\boldsymbol \Lambda {\boldsymbol y_{t - 1}},{\boldsymbol P_{t - 1}}} \right)
	\end{aligned}$
\end{center}
%	\end{equation}
where $\boldsymbol P_{t-1}$ is the variance.
%	which needs to be evaluated in order to obtain the posterior marginal. 

Then, parameters of the posterior distribution $P\left( \boldsymbol Y_K| \boldsymbol X_K, \theta^{old}\right)$ are acquired by backward recursion steps. The procedure is summarized in Algorithm \ref{alg:estep}.
%\ref{alg:estep}.

	\renewcommand{\algorithmicrequire}{\textbf{Inputs:}}
\renewcommand{\algorithmicensure}{\textbf{Outputs:}}
\begin{algorithm}[!tbp] 
	\caption{E-step in PSFA-EWC}\label{alg:estep}
%	\small
	\footnotesize
	\begin{algorithmic}[1]
		\REQUIRE $ \boldsymbol \Sigma_1 $,  $ \boldsymbol \Sigma_x $, $ \boldsymbol \Lambda $,  $\boldsymbol V$, $\boldsymbol X_K$			
		\ENSURE  $ \mathbb E \left[\boldsymbol y_t|\boldsymbol X_K\right] $, $ \mathbb E \left[\boldsymbol y_t \boldsymbol y_{t-1}^T|\boldsymbol X_K\right] $, $ \mathbb E \left[\boldsymbol y_t \boldsymbol y_t^T|\boldsymbol X_K\right] $
		%$ \hat{\boldsymbol \mu}_t $, $ \hat{\boldsymbol U}_t $, $ \hat{\boldsymbol U}_{t,t-1} $, $ {\boldsymbol J}_t $,
		
		\%	 Forward steps by Kalman filter:			
		\STATE Initialize $ 	\boldsymbol K_1 = \boldsymbol \Sigma_1 \boldsymbol V^T \left(\boldsymbol V \boldsymbol \Sigma_1 \boldsymbol V^T + \boldsymbol \Sigma_x\right)^{-1} $, $ \boldsymbol \mu_1 = \boldsymbol K_1 \boldsymbol x_1 $,  $ \boldsymbol U_1 = \left(\boldsymbol I - \boldsymbol K_1 \boldsymbol V \right) \boldsymbol \Sigma_1 $			
		\FOR{$ t=1:T_K $} 			
		\STATE %covariance (prior): 
		$ \boldsymbol P_{t-1} = \boldsymbol \Lambda\left(\boldsymbol U_{t-1} -\boldsymbol I\right)\boldsymbol \Lambda^T+\boldsymbol I $			
		\STATE %Kalman gain:  
		$	\boldsymbol K_t = \boldsymbol P_{t-1} \boldsymbol V^T \left(\boldsymbol V \boldsymbol P_{t-1} \boldsymbol V^T +\boldsymbol \Sigma_x\right)^{-1}$			
		\STATE %updated mean: 
		$ \boldsymbol \mu_t = \boldsymbol \Lambda \boldsymbol \mu_{t-1} +\boldsymbol K_t \left(\boldsymbol x_t -\boldsymbol V \boldsymbol \Lambda \boldsymbol \mu_{t-1}\right) $			
		\STATE %Covariance (posterior): 
		$ \boldsymbol U_t = \left(\boldsymbol I- \boldsymbol K_t \boldsymbol V\right) \boldsymbol P_{t-1} $			
		\ENDFOR	
		
		\%  Backward steps by  RTS smoother  			
		\STATE Initialize $\hat{\boldsymbol \mu}_{T_K} = {\boldsymbol \mu}_{T_K}$, $ \hat{\boldsymbol U}_{T_K} = {\boldsymbol U}_{T_K} $			
		\FOR{$ t=T_K:2 $} 
		\STATE  %Gain: 
		$ \boldsymbol J_{t-1} = \boldsymbol U_{t-1} \boldsymbol \Lambda^T \boldsymbol P_{t-1}^{-1} $			
		\STATE %Mean: 
		$ \hat{\boldsymbol \mu}_{t-1} = {\boldsymbol \mu}_{t-1} + {\boldsymbol J}_{t-1}\left(\hat{\boldsymbol \mu}_{t}- \boldsymbol \Lambda \boldsymbol \mu_{t-1}\right) $			
		\STATE %Covariance: 
		$ \hat{\boldsymbol U}_{t-1} = {\boldsymbol U}_{t-1} +{\boldsymbol J}_{t-1} \left(\hat{\boldsymbol U}_{t} -{\boldsymbol P}_{t-1}\right) {\boldsymbol J}_{t-1}^T $			
		\ENDFOR	
		
		\% Calculate the sufficient statistics			
		\FOR{$ t = 1:T_K $}			
		\STATE $ \mathbb E \left[\boldsymbol y_t|\boldsymbol X_K\right] =  \hat{\boldsymbol \mu}_t $			
		\STATE  $ \mathbb E \left[\boldsymbol y_t \boldsymbol y_{t-1}^T|\boldsymbol X_K\right] =  \boldsymbol J_{t-1} \hat{\boldsymbol U}_t +\hat{\boldsymbol \mu}_t \hat{\boldsymbol \mu}_{t-1}^T $			
		\STATE $ \mathbb E \left[\boldsymbol y_t \boldsymbol y_t^T|\boldsymbol X_K\right] = \hat{\boldsymbol U}_t+\hat{\boldsymbol \mu}_t \hat{\boldsymbol \mu}_{t}^T $			
		\ENDFOR
	\end{algorithmic}
\end{algorithm}

	\bibliography{reference_abrv_v1}
	\bibliographystyle{ieeetr}

\begin{IEEEbiography}[{\includegraphics[width=1in,height=1.25in,clip,keepaspectratio]{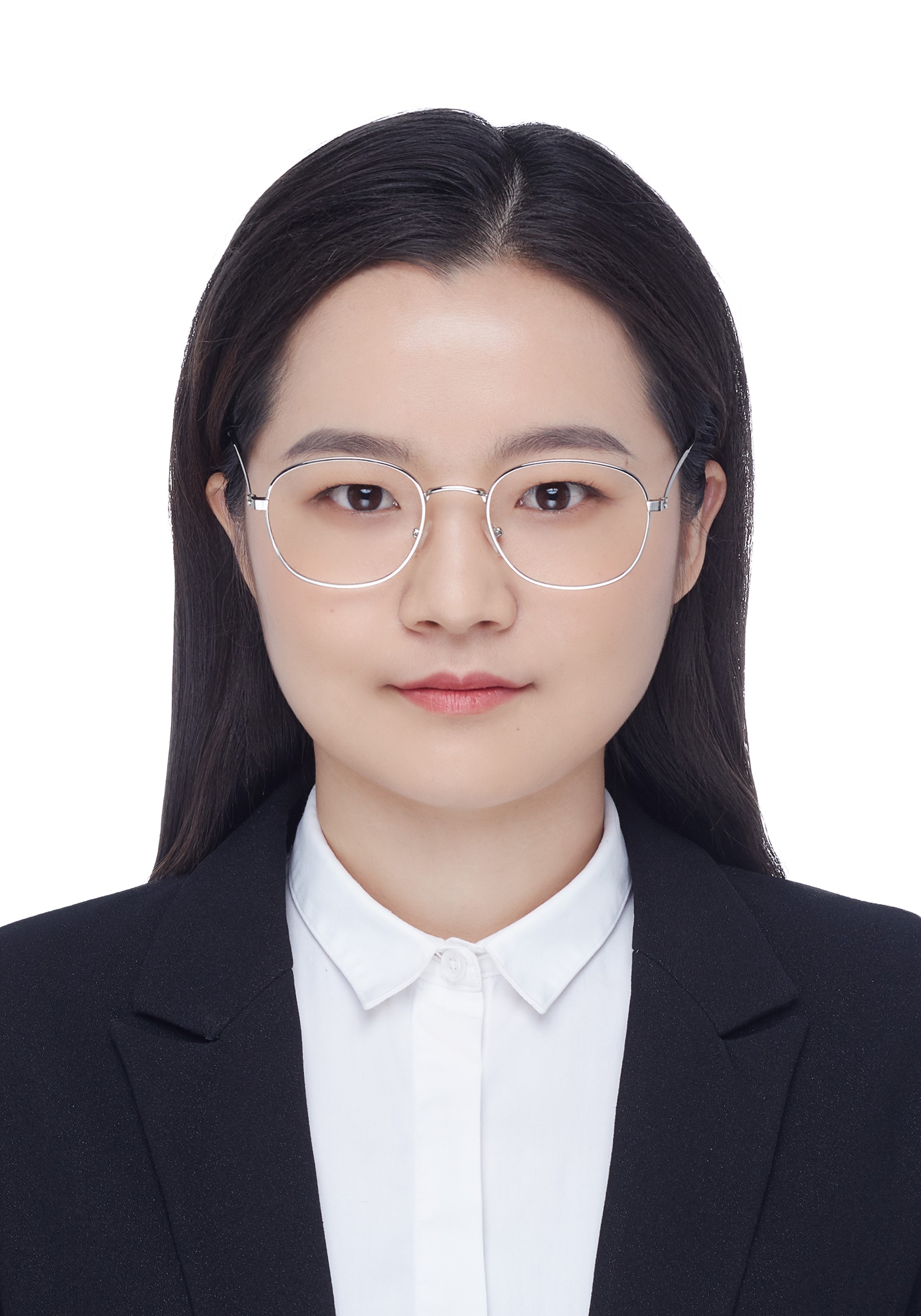}}]
	{Jingxin Zhang}
	received  B.E. degree from School of Electrical Engineering and Automation, Harbin Engineering University, Harbin, China, the M.E. degree from Control Science and Engineering, Harbin Institute of Technology, Harbin, China, in 2014 and 2016, respectively. Since 2018, she has been pursuing the
	Ph.D. degree with the Department of Automation, Tsinghua University, Beijing, China.
	Her research interests are data-driven fault detection and diagnosis, performance monitoring and their applications in the industrial process.
\end{IEEEbiography}

\begin{IEEEbiography}[{\includegraphics[width=1in,height=1.25in,clip,keepaspectratio]{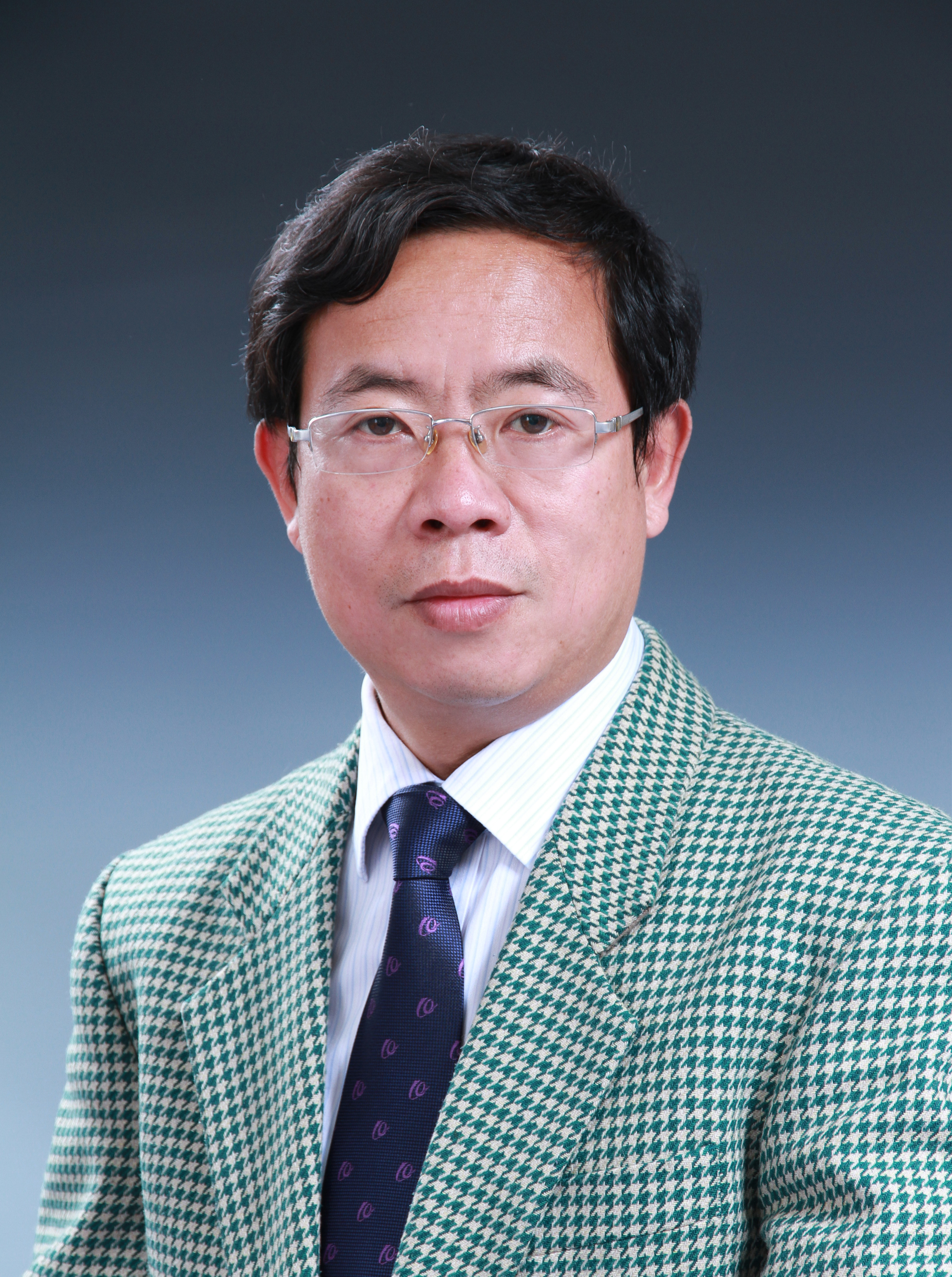}}]{Donghua Zhou}
	(SM'99-F'19, IEEE) received the B.Eng., M. Sci., and Ph.D. degrees all in electrical engineering from Shanghai Jiaotong University, China, in 1985, 1988, and 1990, respectively. He was an Alexander von Humboldt research fellow with the university of Duisburg, Germany from 1995 to 1996, and a visiting scholar with Yale university, USA from 2001 to 2002. He joined Tsinghua university in 1996, and was promoted as full professor in 1997, he was the head of the department of automation, Tsinghua university, during 2008 and 2015. He is now a vice president, Shandong University of Science and Technology, and a joint professor of Tsinghua university. He has authored and coauthored over 220 peer-reviewed international journal papers and 7 monographs in the areas of fault diagnosis, fault-tolerant control and operational safety evaluation. Dr. Zhou is a fellow of IEEE, CAA and IET, a member of IFAC TC on SAFEPROCESS, an associate editor of Journal of Process Control, the vice Chairman of Chinese Association of Automation (CAA) the TC Chair of the SAFEPROCESS committee, CAA. He was also the NOC Chair of the 6th IFAC Symposium on SAFEPROCESS 2006.
\end{IEEEbiography}

\begin{IEEEbiography} [{\includegraphics[width=1in,height=1.25in,clip,keepaspectratio]{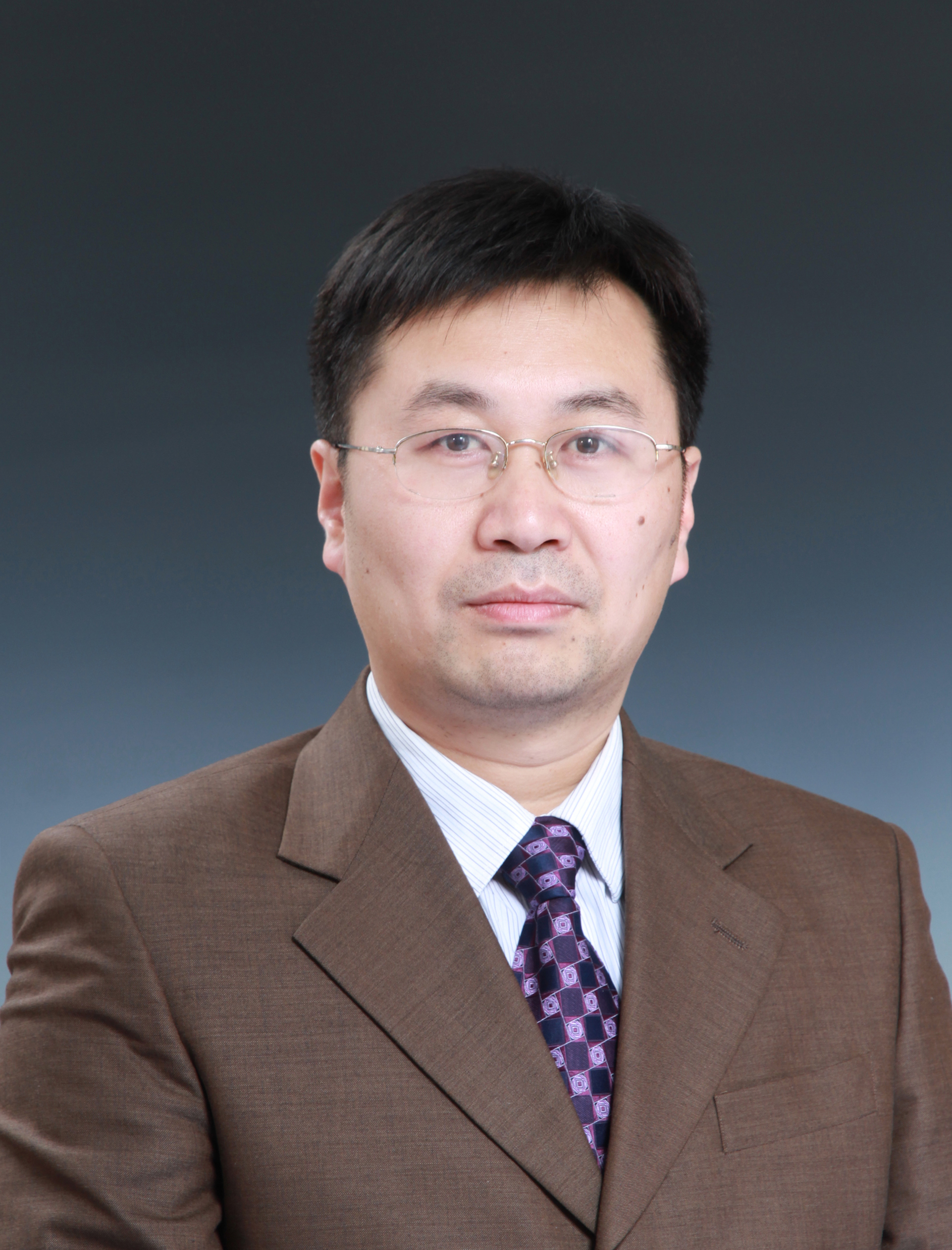}}]
	{Maoyin Chen}
	received the B.S. degree in mathematics and the M.S. degree in control theory and control engineering from Qufu Normal University, Shandong, China, in 1997 and 2000, respectively, and the Ph.D. degree in control theory and control engineering from Shanghai Jiaotong University, Shanghai, China, in 2003.
	From 2003 to 2005, he was a Postdoctoral Researcher with the Department of Automation, Tsinghua University, Beijing, China. From 2006 to 2008, he visited Potsdam University, Potsdam, Germany, as an Alexander von Humboldt Research Fellow. Since October 2008, he has been an Associated Professor with the Department of Automation, Tsinghua University. He has authored and coauthored over 90 peer-reviewed international journal papers. He has won the first prize in natural science (2011, ranked first) and the second prize (2019, ranked first) of CAA. His research interests include fault prognosis and complex systems.
\end{IEEEbiography}

\begin{IEEEbiography}[{\includegraphics[width=1in,height=1.25in,clip,keepaspectratio]{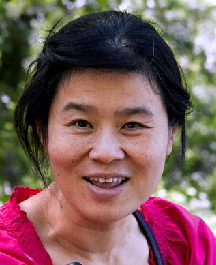}}]{Xia Hong}
	received the B.Sc. and M.Sc. degrees from the National University of Defense Technology, China, in 1984 and 1987, respectively, and the Ph.D. degree from The University of 	Sheffield, U.K., in 1998, all in automatic control. She was a Research Assistant with the Beijing 	Institute of Systems Engineering, Beijing, China, from 1987 to 1993. She was a Research Fellow 	with the Department of Electronics and Computer Science, University of Southampton, from 1997 to 2001.
	
	She is currently a Professor with the Department of Computer Science, School of Mathematical,  Physical and Computational Sciences, University of Reading. She is actively involved in research into nonlinear systems identification, data modeling, estimation and intelligent control,
	neural networks, pattern recognition, learning theory, and their applications. 	She has authored over 170 research papers, and co-authored a research book. 	Dr. Hong received the Donald Julius Groen Prize from IMechE in 1999.
\end{IEEEbiography}

\end{document}